\documentclass[conference]{IEEEtran}
\usepackage{times}

\usepackage[numbers]{natbib}
\usepackage{multicol}
\usepackage[bookmarks=true]{hyperref}
\usepackage{graphicx}  
\usepackage{subcaption}
\usepackage{array}
\usepackage{makecell}
\usepackage{multirow}
\usepackage[export]{adjustbox}
\usepackage{dblfloatfix}
\usepackage{float}

\usepackage{amsmath}
\usepackage{amsfonts}
\usepackage{amssymb}
\usepackage{amsthm}
\usepackage{bm}
\usepackage{bbm}
\usepackage{mathtools}
\usepackage{algorithm}
\usepackage{algorithmic}
\usepackage[utf8]{inputenc}
\usepackage{color}
\usepackage{graphicx}
\usepackage{comment}

\theoremstyle{plain}
\newtheorem{lemma}{Lemma}

\theoremstyle{definition}
\newtheorem{definition}{Definition}

\theoremstyle{remark}

\def\DD{\mathcal{D}}

\def\LL{\mathcal{L}}
\def\MM{\mathcal{M}}\def\NN{\mathcal{N}}

\def\Abb{\mathbb{A}}
\def\Ebb{\mathbb{E}}

\def\Obb{\mathbb{O}}

\def\Sbb{\mathbb{S}}

\newcommand{\reals}{{\bf R}}

\newcommand{\norm}[1]{ \| #1  \|  }

\newcommand{\E}{\Ebb}

\newcommand{\Cov}{\mathrm{Cov}}

\newcommand{\ie}{{\it i.e.}}

\begin{document}

\title{Agile Autonomous Driving using\\ End-to-End Deep Imitation Learning }




%
\author{\authorblockN{Yunpeng Pan\authorrefmark{1},
Ching-An Cheng\authorrefmark{1},
Kamil Saigol\authorrefmark{1}, 
Keuntaek Lee\authorrefmark{2},
Xinyan Yan\authorrefmark{1},\\
Evangelos A. Theodorou\authorrefmark{1}, and
Byron Boots\authorrefmark{1}}
\authorblockA{\authorrefmark{1}Institute for Robotics and Intelligent Machines, 
\authorrefmark{2}School of Electrical and Computer Engineering\\
Georgia Institute of Technology,
Atlanta, Georgia 30332--0250\\ 
\small{ \tt \{ypan37,cacheng,kamilsaigol,keuntaek.lee,xyan43\}@gatech.edu} \\
\small{  \tt evangelos.theodorou@gatech.edu, bboots@cc.gatech.edu}
}
}

\maketitle

\begin{abstract}
We present an end-to-end imitation learning system for agile, off-road autonomous driving using only low-cost on-board sensors. By imitating a model predictive controller equipped with advanced sensors, we train a deep neural network control policy to map raw, high-dimensional observations to continuous steering and throttle commands. Compared with recent approaches to similar tasks, our method requires neither state estimation nor on-the-fly planning to navigate the vehicle. Our approach relies on, and experimentally validates, recent imitation learning theory. Empirically, we show that  policies trained with online imitation learning overcome well-known challenges related to covariate shift and generalize better than policies trained with batch imitation learning. Built on these insights, our autonomous driving system demonstrates successful high-speed off-road driving, matching the state-of-the-art performance.

\end{abstract}

\IEEEpeerreviewmaketitle

\section{Introduction}
\begin{figure}
	\centering
	\includegraphics[width=0.4\textwidth]{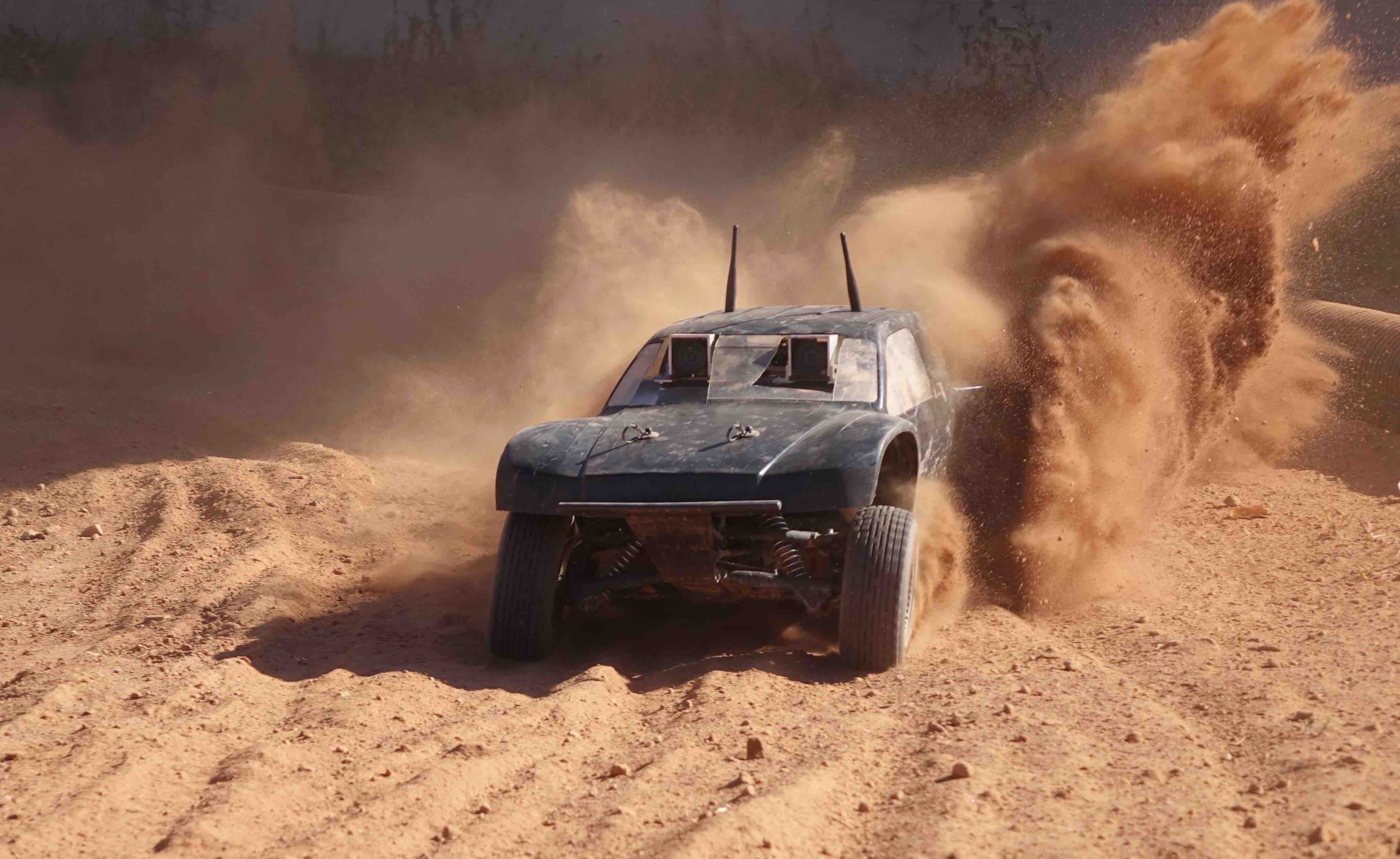}
	\caption{ The high-speed off-road driving task.}
	\label{fig:task}
	\vspace{-6mm}
\end{figure}

High-speed autonomous off-road driving is a challenging robotics problem~\cite{michels2005high,williams2016aggressive,Williams-ICRA-17} (Fig.~\ref{fig:task}). To succeed in this task, a robot is required to perform both precise steering {and} throttle maneuvers in a physically-complex, uncertain environment by executing a series of high-frequency decisions. Compared with most previously studied autonomous driving tasks, the robot here must reason about minimally-structured, stochastic natural environments and operate at high speed. Consequently, designing a control policy by following the traditional model-plan-then-act approach~\cite{michels2005high,paden2016survey} becomes challenging, as it is difficult to adequately characterize the robot's interaction with the environment \emph{a priori}. 


This task has been considered previously, for example, by Williams et al.~\cite{williams2016aggressive,Williams-ICRA-17} using model-predictive control (MPC). While the authors demonstrate impressive results, their internal control scheme relies on expensive and accurate Global Positioning System (GPS) and Inertial Measurement Unit (IMU) for state estimation and demands high-frequency online replanning for generating control commands. Due to these costly hardware requirements, their robot can only operate in a rather controlled environment, which limits the applicability of their approach. 

We aim to relax these requirements by designing a reflexive driving policy that uses only \emph{low-cost, on-board} sensors (e.g. monocular camera, wheel speed sensors). Building on the success of deep reinforcement learning (RL)~\cite{Levine:2016,mnih2015human},
we adopt deep neural networks (DNNs) to parametrize the control policy and learn the desired parameters from the robot's interaction with its environment. 
While the use of DNNs as policy representations for RL is not uncommon, in contrast to most previous work that showcases RL in simulated environments~\cite{mnih2015human}, our agent is a high-speed physical system that incurs real-world cost: collecting data is a cumbersome process, and a single poor decision can physically impair the robot and result in weeks of time lost while replacing parts and repairing the platform. Therefore, direct application of model-free RL techniques is not only sample inefficient, but costly and dangerous in our experiments.

\begin{table*}[t]
	\caption{Comparison of our method to prior work on IL for autonomous driving}\label{tab:comparison of IL} 
	\vspace{-4mm}
	\begin{center} 
		{\begin{tabular}{|c|c|c|c|c|c|c| } 
				\hline
				Methods & Tasks & Observations   &  \makecell{Action} & \makecell{Algorithm} &  Expert  & Experiment\\ 
				\hline\hline
				\cite{bojarski2016end} & On-road low-speed & Single  image & Steering & Batch &  Human & Real  $\&$simulated   \\ 
				\hline
				\cite{pomerleau1989alvinn}\ & On-road low-speed & Single  image $\&$ laser & Steering & Batch & Human & Real  $\&$simulated   \\ 
				\hline
				\cite{Viktor2017}  & On-road low-speed & Single  image & Steering & Batch & Human & Simulated  \\ 
				\hline
				\cite{muller2006off}  & Off-road low-speed & Left $\&$ right   images & Steering & Batch & Human& Real\\
				\hline
				\cite{zhang2016query}  & On-road  unknown speed & Single  image & Steering + break & Online &  Pre-specified policy & Simulated \\
				\hline 
				\bf \makecell{Our \\Method} & Off-road high-speed   &  \makecell{Single  image + \\wheel speeds} & Steering + throttle  & \makecell{Batch $\&$\\ online}& Model predictive controller& \makecell{Real $\&$\\ simulated}\\ 
				\hline
		\end{tabular}}
	\end{center}
	\vspace{-7mm}
\end{table*}

These real-world factors motivate us to adopt  \emph{imitation learning} (IL)~\cite{pomerleau1989alvinn} to optimize the control policy instead. A major benefit of using IL is that we can leverage domain knowledge through \emph{expert} demonstrations. 
This is particularly convenient, for example, when there already exists an autonomous driving platform built through classic system engineering principles. While such a system (e.g.~\cite{williams2016aggressive}) 
usually requires expensive sensors and dedicated computational resources, with IL we can train a lower-cost robot to behave similarly, without carrying the expert's hardware burdens over to the learner. Here we assume the expert is given as a black box oracle that can provide the desired actions when queried, as opposed to the case considered in~\cite{kahn2017plato} where the expert can be modified to accommodate the learning progress.

In this work, we present an IL system for real-world high-speed off-road driving tasks. \footnote{\href{https://www.youtube.com/channel/UCc8KdLVDMEwgCryrfwrL_iA}{Test run videos.}}   By leveraging demonstrations from an algorithmic expert, our system can learn a driving policy that achieves similar performance compared to the expert. The system was implemented on a 1/5-scale autonomous AutoRally car \cite{goldfain2018autorally}. In real-world experiments, we show the AutoRally car---without any state estimator or online planning, but with a DNN policy that directly inputs measurements from a low-cost monocular camera and wheel speed sensors---could learn to perform high-speed driving at an average speed of $\sim$6 m/s and a top speed of $\sim$8 m/s (equivalently  108 km/h and 144 km/h on a full-scale car), matching the state-of-the-art~\cite{Williams-ICRA-17}. 

\section{Related Work}

End-to-end learning for self-driving cars has been explored since the late 1980s. The Autonomous Land Vehicle in a Neural Network (ALVINN)~\cite{pomerleau1989alvinn} was developed to learn steering angles directly from camera and laser range  measurements using a neural network with a single hidden layer. Based on similar ideas, modern self-driving cars~\cite{muller2006off, bojarski2016end,Viktor2017} have recently started to employ a batch IL approach: with DNN control policies, these systems require only expert demonstrations during the training phase and on-board measurements during the testing phase. For example, Nvidia's PilotNet~\cite{bojarski2016end,bojarski2017explaining}, a convolutional neural network that outputs steering angle given an image, was trained to mimic human drivers' reactions to visual input with demonstrations collected in real-world road tests. 

Our problem  differs substantially from these previous on-road driving tasks. We study autonomous driving on a fixed set of dirt tracks, whereas on-road driving must perform well in a larger domain and contend with moving objects such as cars and pedestrians. While on-road driving in urban environments may seem more difficult, our agent must overcome challenges of a different nature. It is required to drive at high speed, on dirt tracks, the surface of which is constantly evolving and highly stochastic. 
As a result, 
high-frequency application of both steering and throttle commands are required in our task, whereas previous work only focuses on steering commands~\cite{muller2006off,bojarski2017explaining,Viktor2017}.  
A Dataset Aggregation (DAgger)~\cite{ross2011reduction} related online IL algorithm for autonomous driving was recently demonstrated in \cite{zhang2016query}, but only considered simulated environments.  A comparison of IL approaches to autonomous driving is presented in Table~\ref{tab:comparison of IL}. 

Our task is  similar to the task considered by Williams et al.~\cite{williams2016aggressive,Williams-ICRA-17} and Drews et al.~\cite{drews2017aggressive}. 
Compared with a DNN policy, their MPC approach has several drawbacks: computationally expensive optimization for planning is required to be performed online at high-frequency, which becomes repetitive for navigating the vehicle on a track after a few laps. 
In~\cite{williams2016aggressive,Williams-ICRA-17}, accurate GPS and IMU feedbacks are also required for state estimation, which may not contain sufficient information to contend with the changing environment in off-road driving tasks.  While the requirement on GPS and IMU is relaxed by using a vision-based cost map in~\cite{drews2017aggressive}, a large dataset (300,000 images) was used to train the model, expensive on-the-fly planning is still required, and speed performance is compromised.
 In contrast to  previous work, our approach off-loads the hardware requirements to an expert. While the expert may use high-quality sensors and more computational power, our agent only needs access to cheap sensors and its control policy can run reactively in high frequency, without on-the-fly planning. Additionally, our experimental results match those in~\cite{williams2016aggressive}, and are faster and more data efficient than that  in~\cite{drews2017aggressive}.

\section{Imitation Learning for Autonomous Driving}

In this section, we give a concise introduction to IL, and discuss the strengths and weakness of deploying a batch or  an online IL algorithm to our task. 
 Our presentation is motivated
by the realization that the  connection between online IL and DAgger-like algorithms~\cite{ross2011reduction} has not been formally
introduced in continuous domains. To our knowledge, DAgger has only been used heuristically in these domains~\citep{ross2013learning,zhang2016query}. Here we simplify the derivation of~\cite{ross2011reduction} and extend it to continuous
action spaces as required in the autonomous driving task.

\subsection{Problem Definition}
To mathematically formulate the autonomous driving task, we consider a discrete-time continuous-valued RL problem. Let $\Sbb$, $\Abb$, and $\Obb$ be the state, action, and the observation spaces. In our setting, the state space is unknown to the agent; observations consist of on-board measurements, including a monocular RGB image from the front-view camera and wheel speeds from Hall effect sensors; actions consist of continuous-valued steering and throttle commands. 

The goal is to find a stationary, reactive policy\footnote{While we focus on reactive policies in this section, the same derivations apply to history-dependent policies.} $\pi
: \Obb \mapsto \Abb$ (e.g. a DNN policy) such that $\pi$ achieves low accumulated cost over a finite horizon of length $T$, 
\begin{equation}
\textstyle
\min_{\pi} J(\pi), \quad J(\pi) \coloneqq  \Ebb_{\rho_\pi} \left[ \sum_{t=0}^{T-1}  c(s_t, a_t)  \right ],   \label{eq:RL learning}
\end{equation}
in which $s_t \in \Sbb$,  $o_t \in \Obb$,   $a_t \in \Abb$,  and 
$\rho_\pi$ is the distribution of trajectory $(s_0, o_0, a_0, s_1, \dots, a_{T-1})$
under policy $\pi$.  
Here $c$ is the instantaneous cost, which, e.g., encourages high speed driving while staying on the track. 
For notation: given a policy $\pi$, we denote $\pi_o$ as the distribution of actions given observation $o$, and $a = \pi(o)$ as the (stochastic) action taken by the policy. 
We denote $Q_{\pi}^t(s,a)$ as the Q-function at state $s$ and time $t$, and $V_{\pi}^t(s) = \E_{a \sim \pi_s}[Q_{\pi}^t(s,a)]$ as its associated value function, where $\E_{a \sim \pi_s}$ is a shorthand of $\E_{o | s} \E_{ a \sim \pi_o }$ denoting the expectation of the action marginal given state $s$.  

\subsection{Imitation Learning} \label{sec:IL}
Directly optimizing~\eqref{eq:RL learning} is challenging for high-speed off-road autonomous driving. Since our task involves a physical robot, model-free RL techniques are intolerably sample inefficient and have the risk of permanently damaging the car when applying a partially-optimized policy in exploration.  Although model-based RL may require fewer samples, it can lead to suboptimal, potentially  unstable results, because it is difficult for a model that uses only on-board measurements to fully capture the complex dynamics of off-road driving. 

Considering these limitations, we propose to solve for policy $\pi$ by IL.  We assume the access to an oracle policy or \emph{expert} $\pi^*$ to generate demonstrations during the training phase. This expert can rely on resources that are unavailable in the testing phase, like additional sensors and computation. For example, the expert can be a computationally intensive optimal controller that relies on exteroceptive sensors (e.g. GPS for state estimation), or an experienced human driver. 

The goal of IL is to perform as well as the expert with an error that has at most linear dependency on $T$. Formally, we introduce a lemma due to Kakade and Langford~\cite{kakade2002approximately} and define what we mean by an \emph{expert}.
\begin{lemma} \label{lm:performance difference}
	Define $d_\pi(s,t) = \frac{1}{T} d_{\pi}^t(s)$ as 
	a generalized stationary time-state distribution, where $d_{\pi}^t$ is the  distribution of state at time $t$ when running policy $\pi$. Let $\pi$ and $\pi'$ be two policies. Then 
	\begin{align}
	J(\pi) = J(\pi') + \E_{s,t \sim d_\pi} \E_{a \sim \pi_s} [ A^t_{\pi'}(s,a)  ]
	\end{align}
	where $ A_{\pi'}^t(s, a) = Q_{\pi'}^t(s,a) - V_{\pi'}^t(s)$ is the (dis)advantage function at time $t$ with respect to running $\pi'$.
\end{lemma}
\begin{definition}\label{def:expert}
	A policy $\pi^*$ is called an \emph{expert} to problem~\eqref{eq:RL learning} if $C_{\pi^*} = \sup_{t \in [0,T-1], s\in \Sbb} \text{Lip}\left(Q^t_{\pi^*}(s,\cdot)\right) \in O(1)$ independent of $T$, where $\text{Lip}(f(\cdot))$ denotes the Lipschitz constant of function $f$ and  $Q^t_{\pi^*}$ is the Q-function at time $t$ of running policy $\pi^*$.
\end{definition}
Because $Q^t_{\pi^*}(s,a)$ is the accumulated cost of taking some action $a$ at time $t$ and then executing the expert policy $\pi^*$ afterwards, the idea behind Definition~\ref{def:expert} is that a reasonable expert policy $\pi^*$ should perform stably under arbitrary action perturbation,  
regardless of where it starts.\footnote{We define the expert here using an uniform Lipschitz constant because the action space in our task is continuous; for discrete action spaces, $\text{Lip}\left(Q^t_{\pi^*}(s,\cdot)\right)$ can be replaced by $\sup_{a\in \Abb} Q_{\pi^*}^t(s,a)$ and the rest applies.} 
As we will see in Section~\ref{sec:comparison}, this requirement provides guidance for whether to choose batch learning vs. online learning to train a policy by imitation.

\subsubsection{Online Imitation Learning} \label{sec:online}
We now present the objective function for the online learning~\cite{shalev2012online} approach to IL. 
Assume $\pi^*$ is an expert to~\eqref{eq:RL learning} and suppose $\Abb$ is a normed space with norm $\norm{\cdot}$.  Let $D_W(\cdot, \cdot)$ denote the Wasserstein metric~\cite{gibbs2002choosing}: for two probability distributions $p$ and $q$ defined on a metric space $\MM$ with metric $d$,
\begin{align}
D_W(p, q) &\coloneqq \sup_{f:\text{Lip}(f(\cdot)) \leq 1 }  \E_{x\sim p}[f(x)] - \E_{x\sim q} [f(x)]  \label{eq:wasserstein def} \\
&= \inf_{\gamma \in \Gamma(p, q)} \int_{\MM \times \MM}d(x,y) d\gamma(x,y), \label{eq:wasserstein def2}
\end{align}
where $\Gamma$ denotes the family of distributions whose marginals are $p$ and $q$. It can be shown by the Kantorovich-Rubinstein theorem that the above two definitions are equivalent~\cite{gibbs2002choosing}. 
These assumptions allow  us to construct a surrogate problem, which is relatively easier to solve than~\eqref{eq:RL learning}. We achieve this by  upper-bounding the difference between the performance of $\pi$ and $\pi'$ given in Lemma~\ref{lm:performance difference}:
\begin{align}
&J(\pi) - J(\pi^*) \nonumber  \\ 
&= \E_{s,t \sim d_\pi}\left[ \E_{a \sim \pi_s } [ Q^t_{\pi^*}(s,a)  ] -\E_{a^* \sim \pi^*_s } [ Q^t_{\pi^*}(s, a^*)  ]\right]  \nonumber \\
&\leq C_{\pi^*}  \E_{s,t \sim d_\pi} \left[  D_{W}(\pi, \pi^*)  \right]  \nonumber \\
&\leq C_{\pi^*}  \E_{s,t \sim d_\pi} \E_{a \sim \pi_s }    \E_{a^* \sim \pi^*_s } [\norm{a - a^*}], \label{eq:online learning upper bound}
\end{align} 
where we invoke the definition of advantage function $A^t_{\pi^*}(s,a) = Q^t_{\pi^*}(s,a) - \E_{a^* \sim \pi^*_s }[Q^t_{\pi^*}(s,a^*)]$, and  the first and the second inequalities are due to~\eqref{eq:wasserstein def} and~\eqref{eq:wasserstein def2}, respectively.

Define $\hat{c}(s,a) = \E_{a^* \sim \pi^*_s } [\norm{a - a^*}] $. Thus, to make $\pi$ perform as well as $\pi^*$, we can minimize the upper bound, which is equivalent to solving a surrogate RL problem
\begin{equation}
\textstyle
\min_{\pi} \Ebb_{\rho_\pi} \left[ \sum_{t=1}^{T}  \hat{c}(s_t, a_t)  \right]. \label{eq:online learning}
\end{equation}
The problem in~\eqref{eq:online learning} is called the \emph{online} IL problem. This surrogate problem is comparatively more structured~ than the original RL problem~\eqref{eq:RL learning}~\citep{cheng2018fast}, so we can adopt algorithms with provable performance guarantees. In this paper, we use the meta-algorithm DAgger~\cite{ross2011reduction}, which reduces~\eqref{eq:online learning} to a sequence of supervised learning problems:
Let $\DD$ be the training data. DAgger initializes $\DD$ with samples gathered by running $\pi^*$. Then, in the $i$th iteration, it trains $\pi_i$ by supervised learning, 
\begin{equation} \textstyle
\pi_i  = \arg\min_{\pi} \Ebb_{\DD}[ \hat{c}(s_t, a_t )  ], \label{eq:supervised learning}  
\end{equation}
where subscript $\DD$ denotes empirical data distribution.
Next it runs $\pi_i$ to collect more data, which is then added  into $\DD$ to train $\pi_{i+1}$. The procedure is repeated for $O(T)$ iterations and the best policy, in terms of~\eqref{eq:online learning}, is returned. Suppose the policy is linearly parametrized. 
Since our instantaneous cost $\hat{c}(s_t, \cdot)$ is strongly convex, 
the theoretical analysis of DAgger applies. Therefore, together with the assumption that $\pi^*$ is an expert, running DAgger to solve~\eqref{eq:online learning} finds a policy $\pi$ with performance  $J(\pi) \leq J(\pi^*) + O(T)$, achieving our initial goal. 

We note here the instantaneous cost $\hat{c}(s_t, \cdot)$ can be selected to be any suitable norm according the problem's property. In our off-road autonomous driving task, we find $l_1$-norm is preferable (e.g. over $l_2$-norm) for its ability to filter outliers in a highly stochastic environment.  


\subsubsection{Batch Imitation Learning} \label{sec:batch}
By swapping the order of $\pi$ and $\pi^*$ in the above derivation in ~\eqref{eq:online learning upper bound}, we can derive another upper bound and use it to construct another surrogate problem: define $\tilde{c}_{\pi}(s^*, a^*) =  \E_{a \sim \pi_{s^*}} [\norm{a - a^*}]$ and $C_\pi^t(s^*) = \text{Lip}(Q^t_{\pi}(s^*,\cdot))$, then 
\begin{align}
&J(\pi) - J(\pi^*) \nonumber \\
&= \E_{s^*,t \sim d_{\pi^*}}\left[ \E_{a \sim \pi_{s^*}} [ Q^t_{\pi}(s^*,a)  ] - \E_{a^* \sim \pi^*_{s^*}} [ Q^t_{\pi}(s^*,a^*)  ]\right]  \nonumber \\
&\leq  \E_{s^*,t \sim d_{\pi^*}} \E_{a^* \sim \pi^*_{s^*}} \left[ C_{\pi}^t(s^*) \tilde{c}_{\pi}(s^*, a^*) \right].   \label{eq:batch learning upper bound}
\end{align}
where we use again Lemma~\ref{lm:performance difference} for the equality and the property of Wasserstein distance for inequality. The minimization of the upper-bound~\eqref{eq:batch learning upper bound} is called the \emph{batch} IL problem~\cite{Viktor2017,bojarski2017explaining}: 
\begin{align}
\textstyle
\min_\pi \Ebb_{\rho_{\pi^*}} \left[ \sum_{t=1}^{T}  \tilde{c}_{\pi}(s^*_t, a^*_t)  \right ], \label{eq:batch learning}
\end{align}
In contrast to the surrogate problem in online IL~\eqref{eq:online learning}, batch IL  reduces to a supervised learning problem, because the expectation is defined by a fixed policy $\pi^*$.

\subsection{Comparison of Imitation Learning Algorithms} \label{sec:comparison}
Comparing~\eqref{eq:online learning upper bound} and~\eqref{eq:batch learning upper bound}, we observe that in batch IL the Lipschitz constant $C_\pi^t(s^*)$, without $\pi$ being an expert as in Definition~\ref{def:expert}, can be on the order of $T-t$ in the worst case. Therefore, if we take a uniform bound and define $C_\pi = \sup_{t \in [0, T-1], s \in \Sbb} C_{\pi}^t (s) $, we see $C_\pi \in O(T)$. In other words, under the same assumption in online imitation (i.e.~\eqref{eq:batch learning upper bound} is minimized to an error in $O(T)$), the difference between $J(\pi)$ and $J(\pi^*)$ in batch IL actually grows quadratically in $T$ due to error compounding. This problem  manifests especially in  stochastic environments.  
Therefore, in order to achieve the same level of performance as online IL, batch IL requires a more expressive policy class or more demonstration samples. As shown in~\cite{ross2011reduction}, the quadratic bound is tight. 

Therefore, if we can choose an expert policy $\pi^*$ that is stable in the sense of Definition~\ref{def:expert}, then online IL is preferred theoretically. This is satisfied, for example, when the expert policy is an algorithm with certain performance characteristics.  On the contrary, if the expert is human, the assumptions required by online IL become hard to realize in real-road driving tasks. 
This is especially true in off-road driving tasks, where the human driver depends heavily on instant feedback from the car to overcome stochastic disturbances. Therefore, the frame-by-frame labeling approach~\cite{ross2013learning}, for example, can lead to a very counter-intuitive, inefficient data collection process, because the required dynamics information is lost in a single image frame. Overall, when using human demonstrations, online IL can be as bad as batch IL~\cite{laskey2016comparing}, simply due to inconsistencies introduced by human nature. 

\section{The Autonomous Driving System} \label{sec:system}
\begin{figure}[t]
	\centering
	\includegraphics[width=0.49\textwidth]{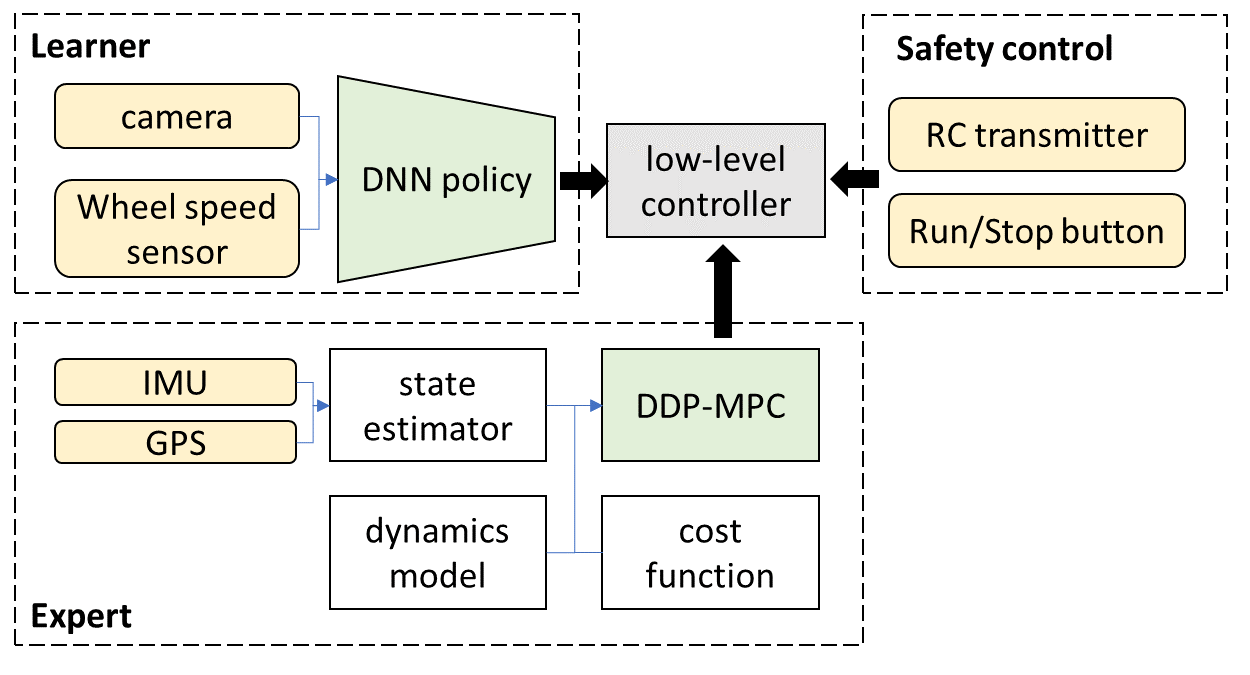}
	\caption{System diagram.}
	\label{fig:system}
	\vspace{-7mm}
\end{figure}
\begin{figure*}[t]
	\vspace{-1mm}
	\centering
	\includegraphics[width=0.85\textwidth]{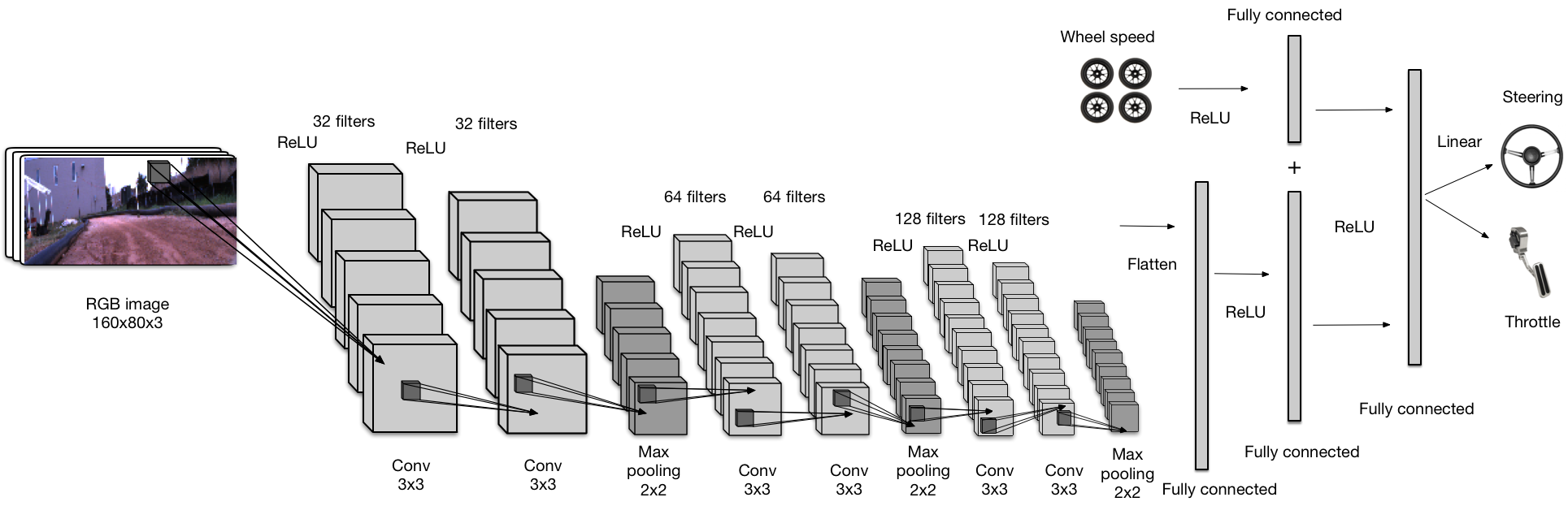}\vspace{-1mm}
	\caption{The DNN control policy.\vspace{-4mm}}
	\label{fig:deep_autorally_net}
	\vspace{-2mm}
\end{figure*}

Building on the previous analyses, we design a system that can learn to perform fast off-road autonomous driving with only on-board measurements.
The overall system architecture for learning end-to-end DNN driving policies is illustrated in Fig.~\ref{fig:system}. It consists of three high-level controllers (an expert, a learner, and a safety control module) and a low-level controller, which receives steering and throttle commands from the high-level controllers and translates them to pulse-width modulation (PWM) signals to drive the steering and throttle actuators of a vehicle. 

On the basis of the analysis in Section~\ref{sec:comparison}, we assume the expert is algorithmic and has access to expensive sensors (GPS and IMU) for accurate global state estimates\footnote{Global position, heading and roll angles, linear velocities, and heading angle rate.} and resourceful computational power. The expert is built on multiple hand-engineered components, including a state estimator, a dynamics model of the vehicle, a cost function of the task, and a trajectory optimization algorithm for planning (see Section~\ref{sec:ddp_expert}). By contrast, the learner is a DNN policy that has access to only a monocular camera and wheel speed sensors and is required to output steering and throttle command directly (see Section~\ref{sec:policy_learning}). In this setting, the sensors that the learner uses can be significantly cheaper than those of the expert; specifically on our experimental platform, the AutoRally car (see Section~\ref{sec:platform}), the IMU and the GPS sensors required by the expert in Section~\ref{sec:ddp_expert} together cost more than \$6,000, while the sensors used by the learner's DNN policy cost less than \$500. 
The safety control module has the highest priority among all three controllers and is used prevent the vehicle from high-speed crashing. 

The software system was developed based on the Robot Operating System (ROS) in Ubuntu. In addition, a Gazebo-based simulation environment \cite{koenig2004use} was built using the same ROS interface but without the safety control module; the simulator was used to evaluate the performance of the software before real track tests.

\vspace{-1mm}
\subsection{An Algorithmic Expert: Model-Predictive Control}\label{sec:ddp_expert}
We use an MPC expert~\cite{pmlr-v70-pan17a} based on an incremental Sparse Spectrum Gaussian Process (SSGP)~\cite{quia2010sparse} dynamics model (which was learned from 30 minute-long driving data) and an iSAM2 state estimator~\cite{kaess2012isam2}.  To generate actions, the MPC expert solves a finite horizon optimal control problem for \textit{every} sampling time: at time $t$, the expert policy  $\pi^*$ is a locally optimal policy such that 
\begin{align} \textstyle
	\pi^* \approx \arg\min_{\pi}   \Ebb_{\rho_\pi} \left[ \sum_{\tau=t}^{t+T_h}  c(s_\tau, a_\tau)  \vert s_t   \right ]  \label{eq:MPC}
\end{align} 
where $T_h$ is the length of horizon it previews. 

The computation is realized by 
Differential Dynamic Programming~(DDP)~\cite{tassa2008receding}: in each iteration of DDP, the system dynamics and the cost function are approximated quadratically along a nominal trajectory; then the Bellman equation of the approximate problem is solved in a backward pass to compute the control law; finally, a new nominal trajectory is generated by applying the updated control law through the dynamics model in a forward pass. 
Upon convergence, DDP returns a locally optimal control sequence $\{\hat{a}_t^*,...,\hat{a}^*_{t+T_h-1}\}$, and the MPC expert executes the first action in the sequence as the expert's action at time $t$ (i.e. $a_t^* = \hat{a}^*_t$). This process is repeated at every sampling time (see the Appendix for details).

In view of the analysis in Section~\ref{sec:IL}, we can assume that the MPC expert satisfies Definition~\ref{def:expert}, because it updates the approximate solution to the original RL problem~\eqref{eq:RL learning} in high-frequency using global state information.
However, because MPC requires replanning for every step, running the expert policy~\eqref{eq:MPC} online consumes significantly more computational power than what is required by the learner. 


\vspace{-1mm}
\subsection{Learning a DNN Control Policy}\label{sec:policy_learning}
The learner's control policy $\pi$ is parametrized by a DNN containing $\sim$10 million parameters. As illustrated in Fig. \ref{fig:deep_autorally_net}, the DNN policy, consists of two sub-networks: a convolutional neural network (CNN) with 6 convolutional layers, 3 max-pooling layers, and 2 fully-connected layers, that takes $160 \times 80$ RGB monocular images as  inputs,\footnote{The raw images from the camera were re-scaled to $160 \times 80$.} and a feedforward network with a fully-connected hidden layer that takes wheel speeds as inputs.  The convolutional and max-pooling layers are used to extract lower-dimensional features from images. The DNN policy uses $3\times 3$ filters for all convolutional layers, and rectified linear unit (ReLU) activation for all layers except the last one. Max-pooling layers with $2\times 2$ filters are integrated to reduce the spatial size of the representation (and therefore reduce the number of parameters and computation loads). The two sub-networks are concatenated and then followed by another fully-connected hidden layer. The structure of this DNN was selected empirically based on experimental studies of several different architectures.

In construction of the surrogate problem for IL, the action space $\Abb$ is equipped with $\norm{\cdot}_1$ for filtering outliers, and the optimization problem, \eqref{eq:supervised learning} or~\eqref{eq:batch learning}, is solved using ADAM~\cite{kingma2014adam}, which is a stochastic gradient descent algorithm with an adaptive learning rate. Note while $s_t$ or $s^*_t$ is used in~\eqref{eq:supervised learning} or~\eqref{eq:batch learning}, the neural network policy does not use the state, but rather the synchronized raw observation $o_t$ as input. Note that we did not perform any data selection or augmentation techniques in any of the experiments. \footnote{Data collection or augmentation techniques such as   \cite{geist2017data,bojarski2016end} can be used in conjunction with our method.} The only pre-processing was scaling and cropping of raw images. 

\vspace{-1mm}
\subsection{The Autonomous Driving Platform}\label{sec:platform}

\begin{figure}[t]
	\centering 
	\includegraphics[width=0.24\textwidth]{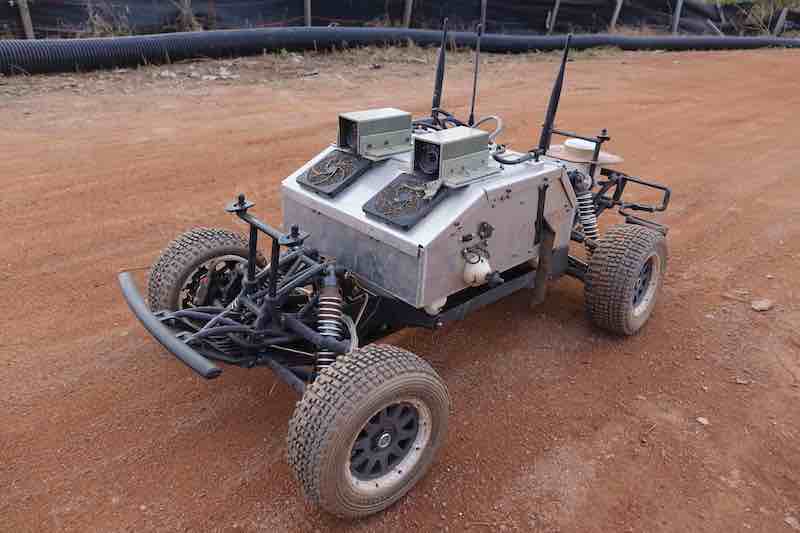}
	\includegraphics[width=0.24\textwidth]{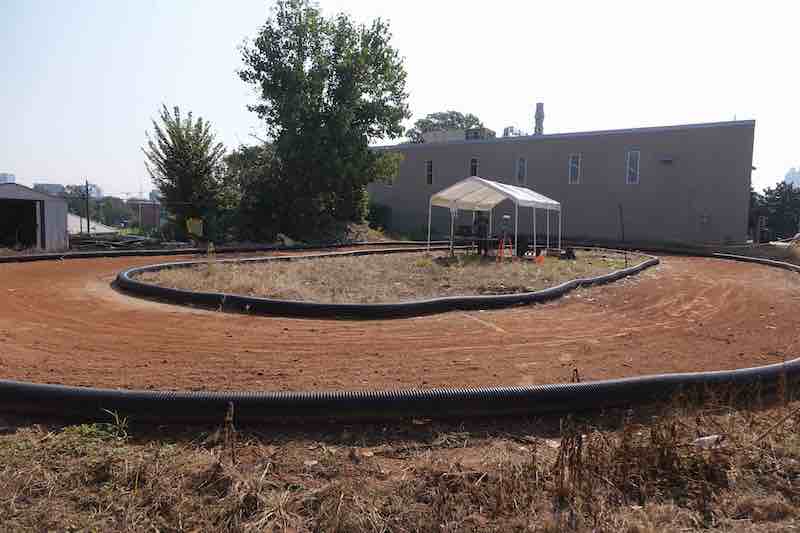}
	\caption{The AutoRally car and the test track.\vspace{-6mm}} \label{fig:autorally}
\end{figure}

To validate our IL approach to off-road autonomous driving, the system was implemented on a custom-built, 1/5-scale autonomous AutoRally car (weight 22 kg; LWH 1m$\times$0.6m$\times$0.4m), shown in the top figure in Fig.~\ref{fig:autorally}. The car was equipped with an ASUS mini-ITX motherboard, an Intel quad-core i7 CPU, 16GB RAM, a Nvidia GTX 750 Ti GPU, and a 11000mAh battery. For sensors, two forward facing machine vision cameras,\footnote{In this work we only used one of the cameras.} a Hemisphere Eclipse P307 GPS module, a Lord Microstrain 3DM-GX4-25 IMU, and Hall effect wheel speed sensors were instrumented. In addition, an RC transmitter could be used to remotely control the vehicle by a human, and a physical run-stop button was installed to disable all motions in case of emergency. The source code used in this work is availalbe \footnote{GitHub repos: \href{https://github.com/ACDSLab/imitation_learning_autorally}{Imitation learning}, \href{http://autorally.github.io/}{AutoRally platform}.}.

In the experiments, all computation was executed on-board the vehicle in real-time. In addition, an external laptop was used to communicate with the on-board computer remotely via Wi-Fi  to monitor the vehicle's status. The observations were sampled and action were executed at 50 Hz to account for the high-speed of the vehicle and the stochasticity of the environment. Note this control frequency is significantly higher than \cite{bojarski2017explaining} (10 Hz), \cite{Viktor2017} (12 Hz), and \cite{muller2006off} (15 Hz).  

\vspace{-1mm}
\section{Experimental Setup}
\label{sec:experiments}
\vspace{-1mm}

\subsection{High-speed Driving Task}
We tested the performance of the proposed IL system in Section~\ref{sec:system} in a high-speed driving task with a desired speed of 7.5 m/s (an equivalent speed of 135 km/h on a full-scale car).
The performance index of the task was formulated as the cost function in the finite-horizon RL problem~\eqref{eq:RL learning} with
\begin{align}
c(s_t,a_t) \hspace{-.25mm}=\hspace{-.25mm} \alpha_1c_\text{pos}(s_t) \hspace{-.25mm}+\hspace{-.25mm}  \alpha_2c_\text{spd}(s_t) 
\hspace{-.25mm} + \hspace{-.25mm}\alpha_3c_\text{slip}(s_t)  \hspace{-.25mm}+  \hspace{-.25mm}\alpha_3c_\text{act}(a_t), \label{eq:cost}
\end{align} \normalsize
in which $c_{\text{pos}}$ favors the vehicle to stay in the middle of the track, $c_{\text{spd}}$ drives the vehicle to reach the desired speed, $c_{\text{slip}}$ stabilizes the car from slipping, and $c_{\text{act}}$  inhibits large control commands (see the Appendix for  details).

The goal of the high-speed driving task to minimize the accumulated cost function over one-minute continuous driving. That is, under the 50-Hz sampling rate, the task horizon was set to 60 seconds ($T= 3000$). The cost information~\eqref{eq:cost} was given to the MPC expert in Fig.~\ref{fig:system} to perform online trajectory optimization with a two-second prediction horizon ($T_h = 100$). In the experiments, the weighting in~\eqref{eq:cost} were set as $\alpha_1=2.5$, $\alpha_2=1$, $\alpha_3=100$ and $\alpha_4=60$, so that the MPC expert in Section~\ref{sec:ddp_expert} could perform reasonably well. The learner's policy was tuned by online/batch IL in attempts to match the expert's performance. 

\vspace{-1mm}
\subsection{Test Track}
All the experiments were performed on an elliptical dirt  track, shown in the bottom figure of Fig.~\ref{fig:autorally}, with the AutoRally car described in Section~\ref{sec:platform}. The test track was $\sim$3m wide and $\sim$30m long and built with fill dirt. Its boundaries were surrounded by soft HDPE tubes, which were detached from the ground, for safety during experimentation. Due to the changing dirt surface, debris from the track's natural surroundings, and the shifting track boundaries after car crashes, the track condition and vehicle dynamics can change from one experiment to the next, adding to the complexity of learning a robust policy.  

\subsection{Data Collection}\label{sec:data_collection}

Training data was collected in two ways. In batch IL, the MPC expert was executed, and the camera images, wheel speed readings, and the corresponding steering and throttle commands were recorded. In online IL, a mixture of the expert and learner's policy was used to collect training data (camera images, wheel speeds, and expert actions): in the $i$th iteration of DAgger, a mixed policy was executed at each time step $ \hat{\pi}_i=\beta^i\pi^* + (1-\beta^i)\pi_{i-1} $, where $\pi_{i-1}$ is learner's DNN policy after $i-1$ DAgger iterations, and $\beta^i$ is the probability of executing the expert policy. The use of a mixture policy was suggested in~\cite{ross2011reduction,cheng2018convergence} for better stability. 
A mixing rate $\beta=0.6$ was used in our experiments. Note that the probability of using the expert decayed exponentially as the number of DAgger iterations increased. 
Experimental data was collected on an outdoor track, and consisted of changing lighting conditions and environmental dynamics. 
In the experiments, the rollouts about to crash were terminated remotely by overwriting the autonomous control commands with the run-stop button or the RC transmitter in the safety control module; these rollouts were excluded from the data collection.

\subsection{Policy Learning}
In online IL, three iterations of DAgger were performed. At each iteration, the robot executed one rollout using the mixed policy described above (the probabilities of executing the expert policy were 60\%, 36\%, and 21\%, respectively). For a fair comparison, the amount of training data collected in batch IL was the same as all of the data collected over the three iterations of online IL.

At each training phase, the optimization problem~\eqref{eq:supervised learning} or~\eqref{eq:batch learning} was solved by ADAM for 20 epochs, with mini-batch size 64, and a learning rate of 0.001. Dropouts were applied at all fully connected layers to avoid over-fitting (with probability 0.5 for the firstly fully connected layer and 0.25 for the rest). See Section \ref{sec:policy_learning} for details. Finally, after the entire learning session of a policy, three rollouts were performed using the learned policy for performance evaluation.

\section{Experimental Results}

\begin{figure}
	\centering
	\begin{subfigure}{0.22\textwidth}
		\centering
		\includegraphics[trim={1cm 8cm 5cm 6cm},clip,width=\textwidth]{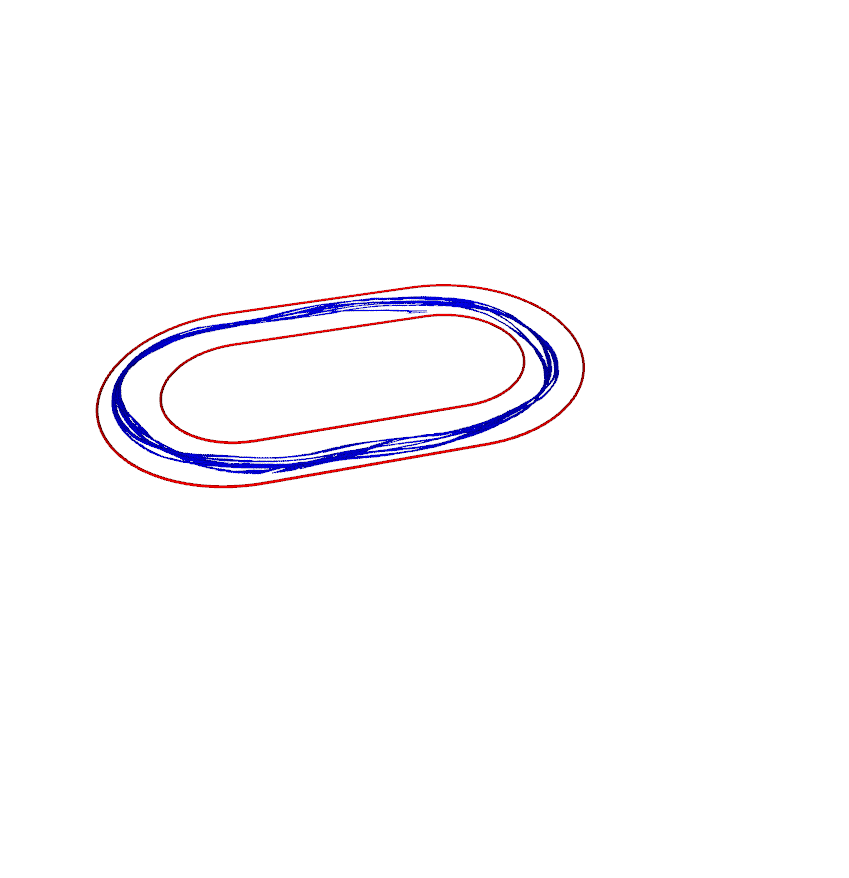}
		\vspace{1mm}\caption{MPC expert. }
	\end{subfigure}
	\begin{subfigure}{0.22\textwidth}
		\centering
		\includegraphics[trim={1cm 7cm 5cm 6cm},clip,width=\textwidth]{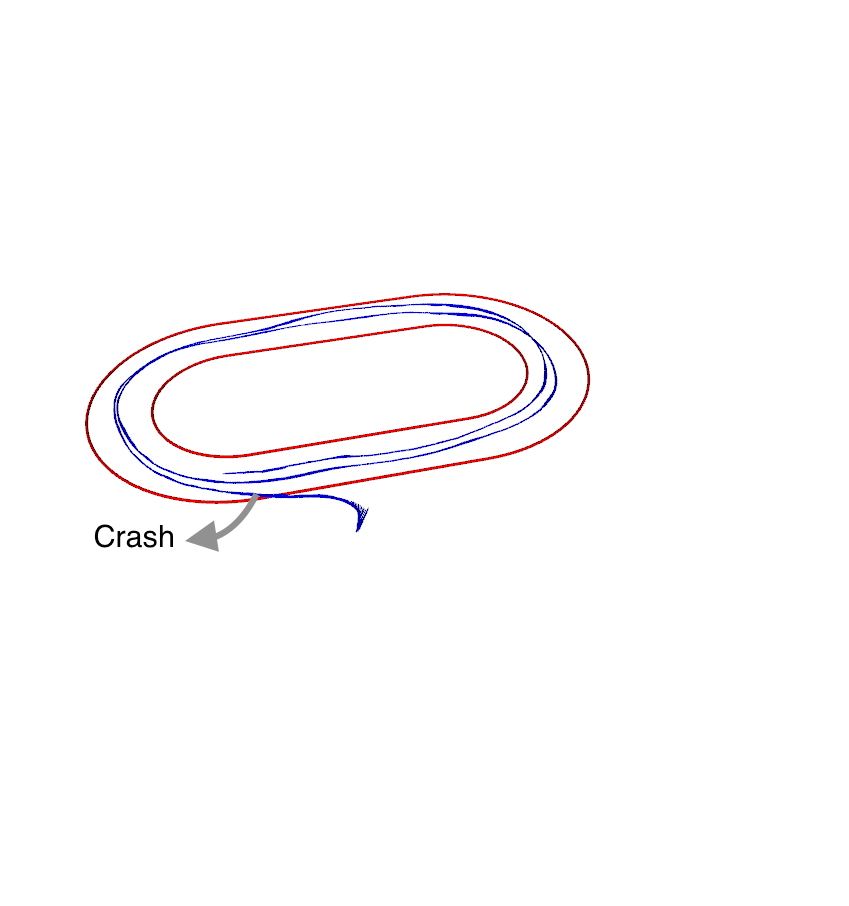}
		\caption{Batch IL. }
	\end{subfigure}
	\begin{subfigure}{0.22\textwidth}
		\centering
		\includegraphics[trim={1cm 7cm 5cm 6cm},clip,width=\textwidth]{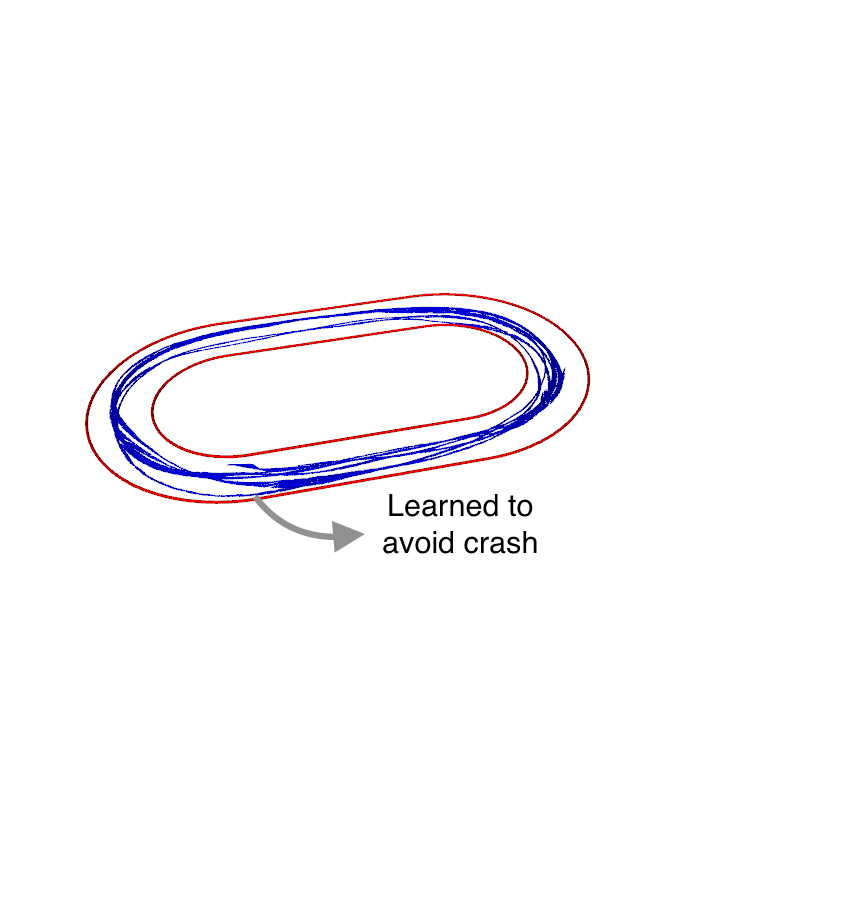}
		\caption{Online IL. }
	\end{subfigure}
	\caption{Examples of vehicle trajectories, where online IL avoids the crashing case encountered by batch IL.  (b) and (c) depict the test runs after training on 9,000 samples.}
	\label{fig:traj}
	\vspace{-3mm}
\end{figure}

\subsection{Empirical Performance}

\begin{table*}
	\caption{Test statistics.
		Total loss denotes the imitation loss in~\eqref{eq:online learning}, which is the average of the steering and the throttle losses. Completion  is defined as the ratio of the traveled time steps to the targeted time steps (3,000). All results here represent the average performance over three independent evaluation trials. 
	 }\label{tab:exp results}
	\begin{center} {\small
			\begin{tabular}{ |c|c|c|c|c|c|c|c|c }   
				\hline
				Policy & Avg. speed & Top speed & Training data  & Completion ratio &  Total loss  & Steering/Throttle loss \\ 
				\hline\hline
				Expert & 6.05 m/s& 8.14 m/s & N/A &  100 $\%$ &   0 & 0  \\ 
				\hline
				Batch& 4.97 m/s & 5.51 m/s & 3000  & 100 $\%$ &   0.108  & 0.092/0.124 \\ 
				
				Batch & 6.02 m/s & 8.18 m/s & 6000 & 51 $\%$ &   0108 & 0.162/0.055  \\ 
				
				Batch & 5.79 m/s& 7.78 m/s  & 9000 &  53 $\%$ &   0.123 & 0.193/0.071 \\ 
				
				Batch & 5.95 m/s& 8.01 m/s & 12000 &  69 $\%$ &   0.105  & 0.125/0.083 \\ 				
				\hline
				Online (1 iter) & 6.02 m/s & 7.88 m/s & 6000 &  100 $\%$ &   0.090 & 0.112/0.067\\ 
				
				Online (2 iter) & 5.89 m/s & 8.02 m/s & 9000  &  100 $\%$ &   0.075  & 0.095/0.055\\ 
				
				Online (3 iter) & 6.07 m/s & 8.06 m/s & 12000  &  100 $\%$ &   0.064  & 0.073/0.055 \\ 
				\hline
				
		\end{tabular}}
	\end{center}
	\vspace{-4mm}
\end{table*}
\vspace{-1mm}

We first study the performance of training a control policy with online and batch IL algorithms. Fig.~\ref{fig:traj} illustrates the vehicle trajectories of different policies. Due to accumulating errors, the policy trained with batch IL crashed into the lower-left boundary, an area of the state-action space rarely explored in the expert's demonstrations. In contrast to batch IL, online IL successfully copes with corner cases as the learned policy occasionally ventured into new areas of the state-action space.

\begin{figure}
	\centering
	\begin{subfigure}[b]{0.33\textwidth}
		\includegraphics[width=\textwidth]{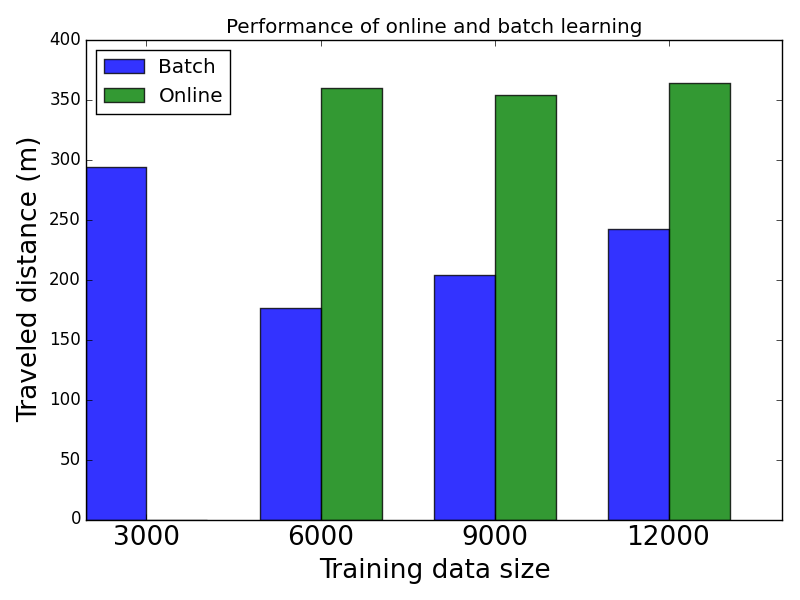}
	\end{subfigure}
	\caption[Performance]{Performance of online and batch IL in the distance (meters) traveled without crashing. The policy trained with a batch of 3,000 samples was used to initialize online IL.}
	\label{fig:online_vs_batch} 
	\vspace{-6mm}
\end{figure}

Fig.~\ref{fig:online_vs_batch} shows the performance in terms of distance traveled without crashing\footnote{We used the safe control module shown in Fig.~\ref{fig:system} to manually terminate the rollout when the car crashed into the soft boundary.} and Table~\ref{tab:exp results} shows the statistics of the experimental results.  Overall, DNN policies trained with both online and batch IL algorithms were able to achieve speeds similar to the MPC expert. However, with the same amount of training data, the policies trained with online IL in general outperformed those trained with batch IL. In particular, the policies trained using online IL achieved better performance in terms of both completion ratio and imitation loss.

In addition, we found that, when using online IL, the performance of the policy monotonically improves over iterations as data are collected, which is opposed to what was found by~\citet{laskey2016comparing}. The discrepancy can be explained with a recent theoretical analysis by~\citet{cheng2018convergence}, which provides a necessary and sufficient condition for the convergence of the policy sequence. In particular, the authors show that adopting a non-zero mixing (as used in our experiment) is sufficient to guarantee the convergence of the learned policy sequence. Our autonomous driving system is a successful real-world demonstration of this IL theory.

Finally, it is worth noting that the traveled distance of the batch learning policy, learned with 3,000 samples, was longer than that of other batch learning policies. This is mainly because this policy achieved better steering performance than throttle performance (cf. Steering/Throttle loss in Table~\ref{tab:exp results}). That is, although the vehicle was able to navigate without crashing, it actually traveled at a much slower speed. By contrast, the other batch learning policies that used more data had better throttle performance and worse steering performance, resulting in faster speeds but also higher chances of crashing.

\subsection{Generalizability of the Learned Policy}\label{sec:generalizability}

\begin{figure}
	\centering
	\begin{subfigure}{0.23\textwidth}
		\centering
		\includegraphics[trim={0.6cm 0.5cm 1.5cm 1.2cm},clip,width=\textwidth]{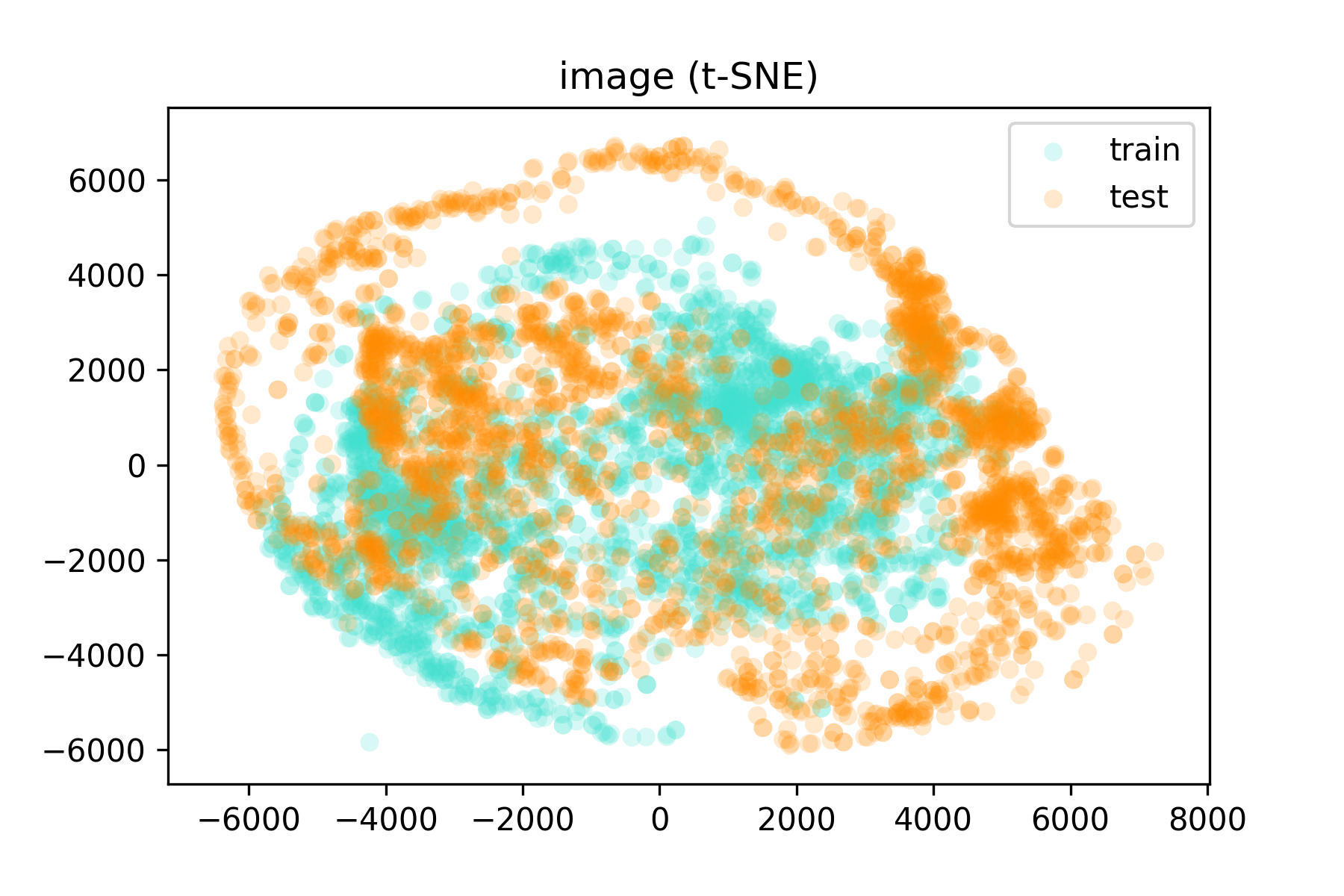}
		\caption{Batch raw image}
	\end{subfigure}
	\begin{subfigure}{0.23\textwidth}
		\centering
		\includegraphics[trim={0.6cm 0.5cm 1.5cm 1.2cm},clip,width=\textwidth]{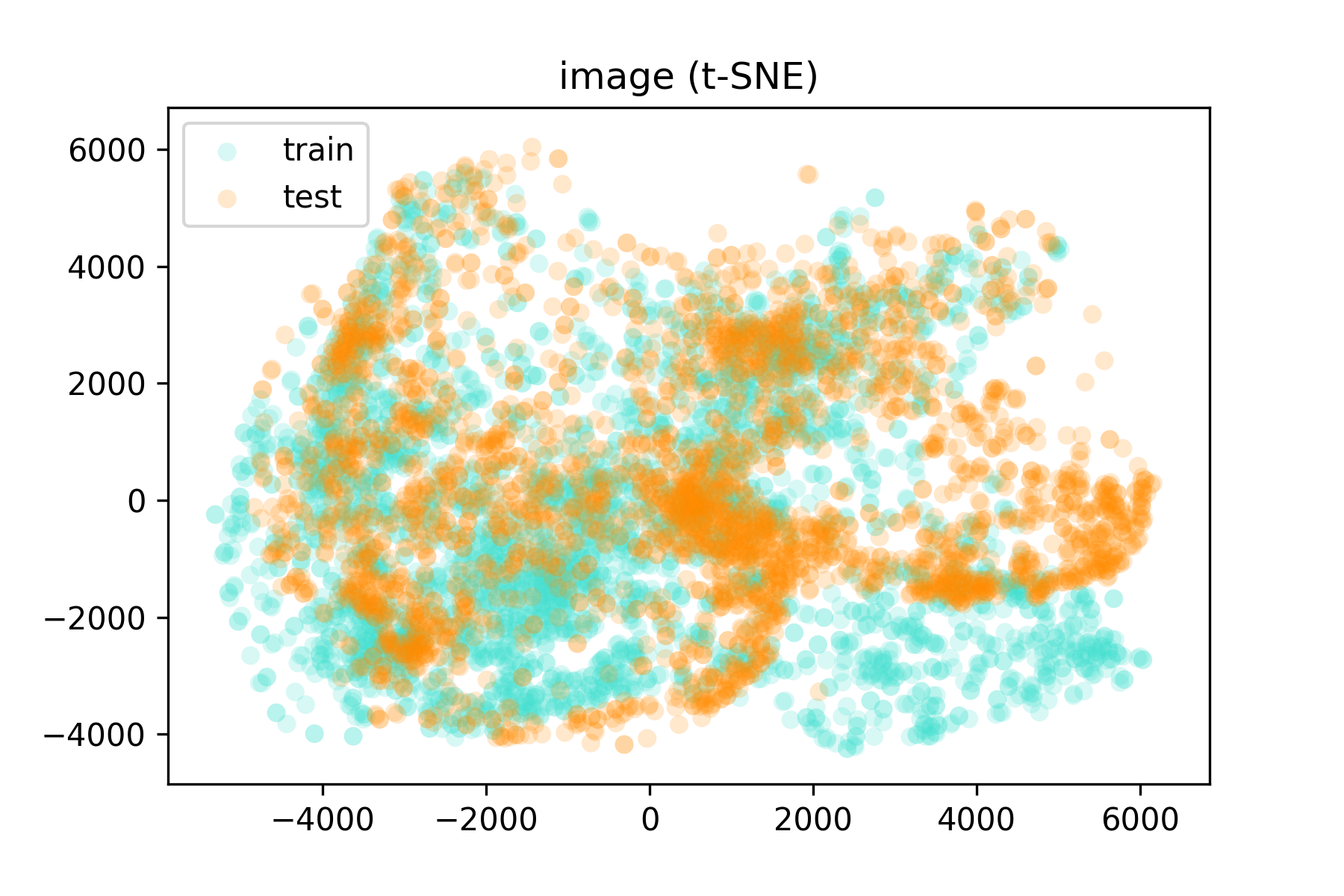}
		\caption{Online raw image}
	\end{subfigure}\\
	\begin{subfigure}{0.23\textwidth}
		\centering
		\includegraphics[trim={1.cm 0.5cm 1.5cm 1.2cm},clip,width=\textwidth]{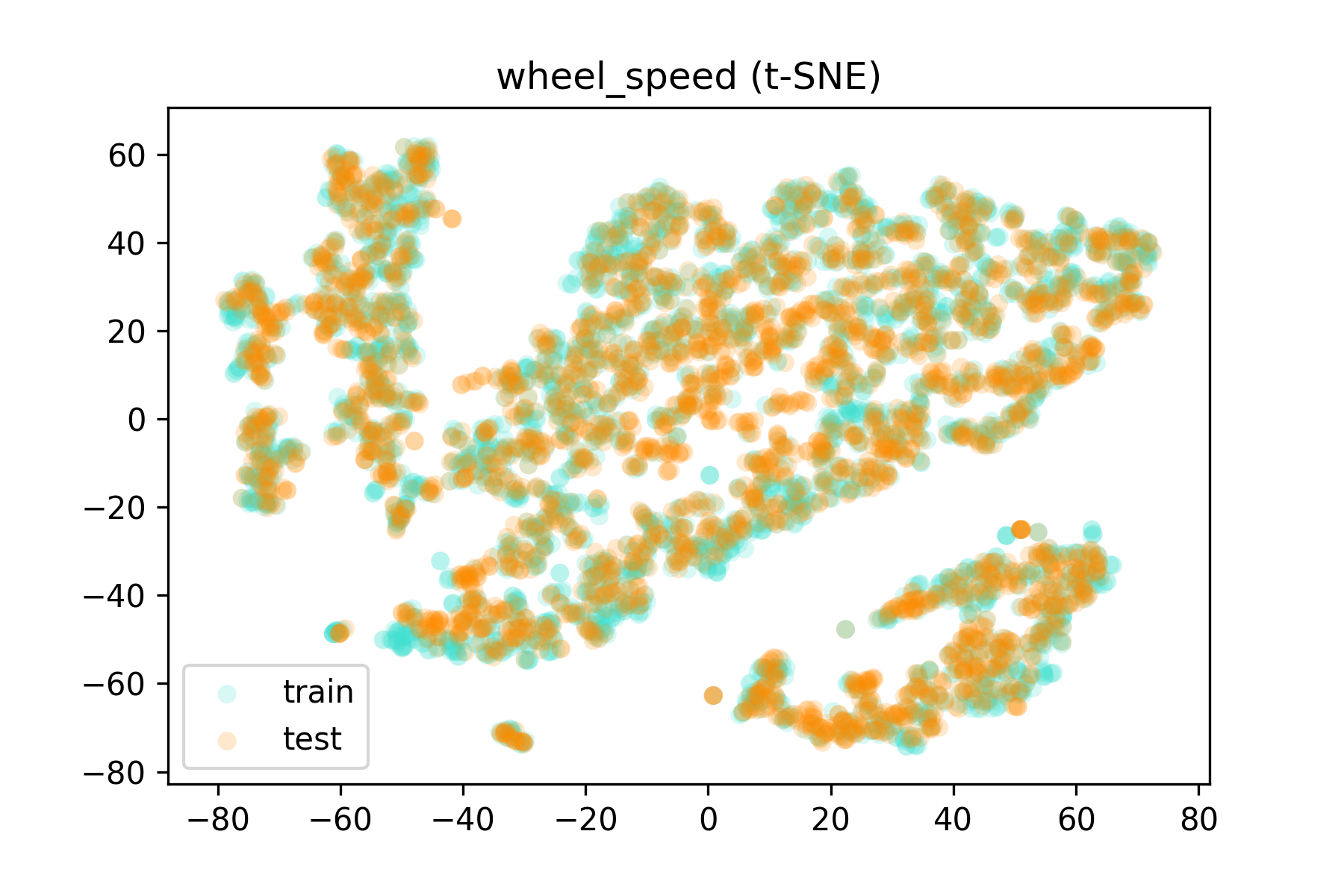}
		\caption{Batch wheel speed}
	\end{subfigure}
	\begin{subfigure}{0.23\textwidth}
		\centering
		\includegraphics[trim={1.cm 0.5cm 1.5cm 1.2cm},clip,width=\textwidth]{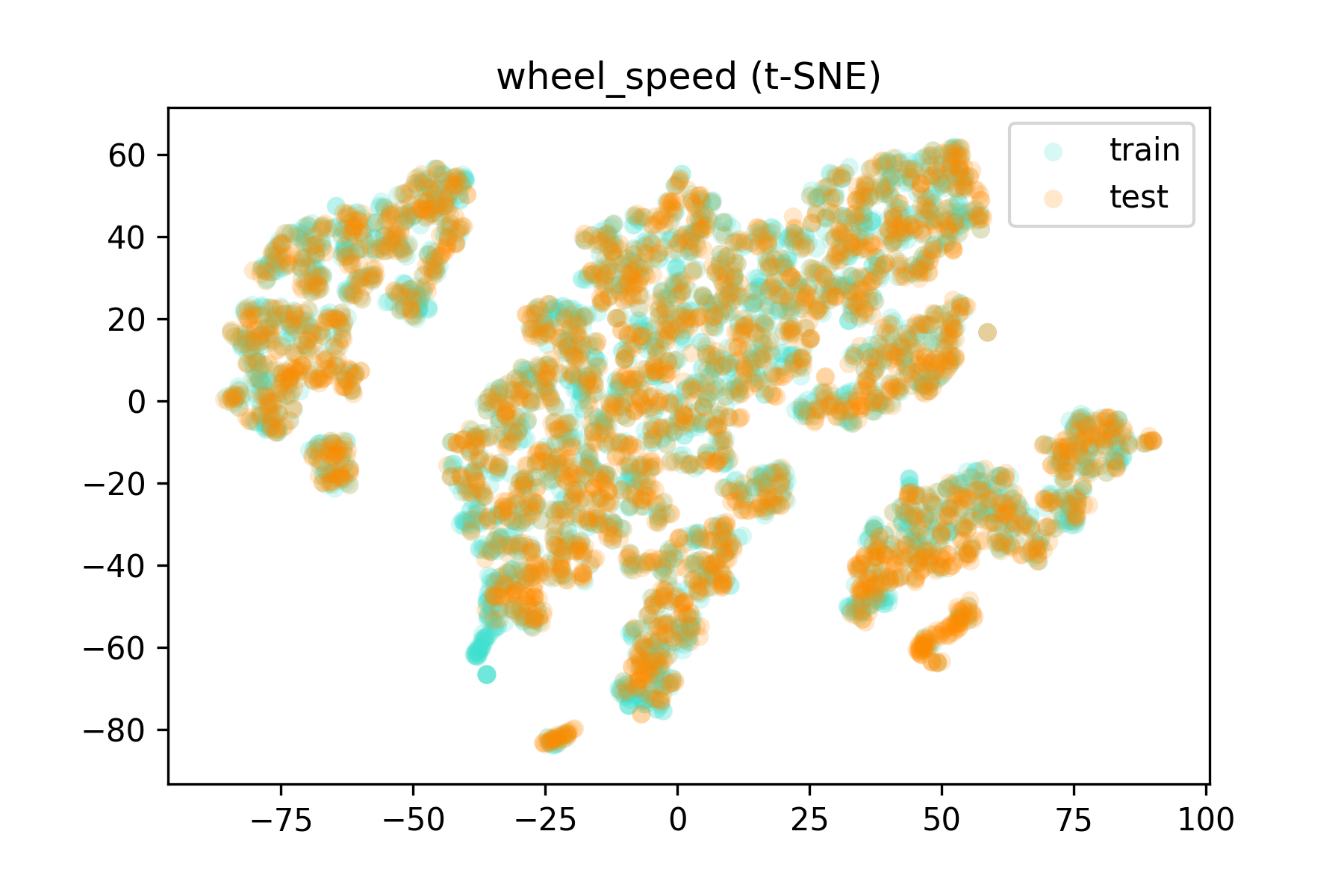}
		\caption{Online wheel speed}
	\end{subfigure}
	\caption{The distributions (t-SNE) of the raw images and wheel speed used as DNN policy's inputs (details in Section~\ref{sec:generalizability}).}
	\label{fig:raw data distributions}
	\vspace{-3mm}
\end{figure}

\begin{figure}
	\centering
	\begin{subfigure}{0.23\textwidth}
		\centering
		\includegraphics[trim={1.cm 0.5cm 1.5cm 1.2cm},clip,width=\textwidth]{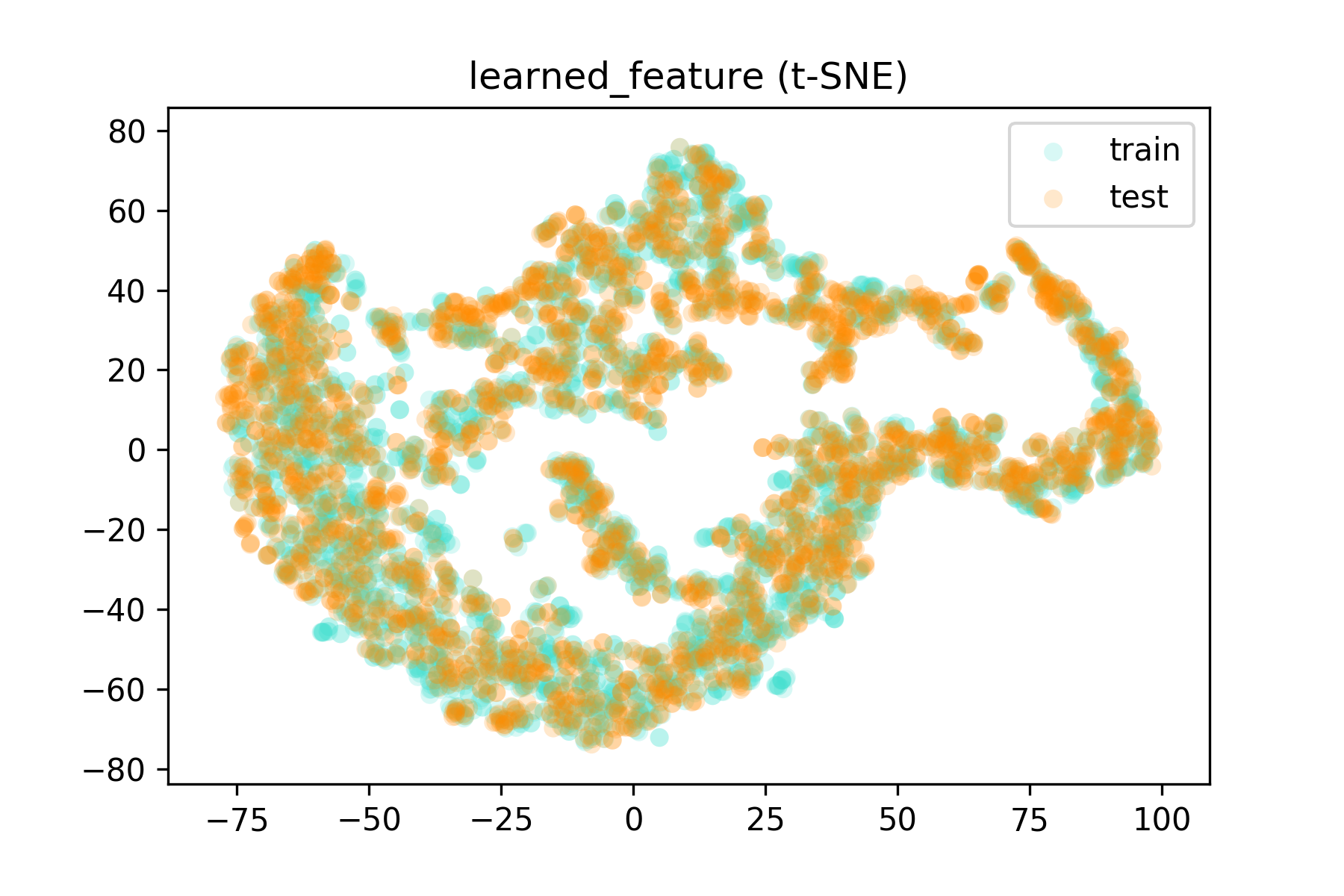}
		\caption{Batch data wrt online model}
	\end{subfigure}
	\begin{subfigure}{0.23\textwidth}
		\centering
		\includegraphics[trim={1.cm 0.5cm 1.5cm 1.2cm},clip,width=\textwidth]{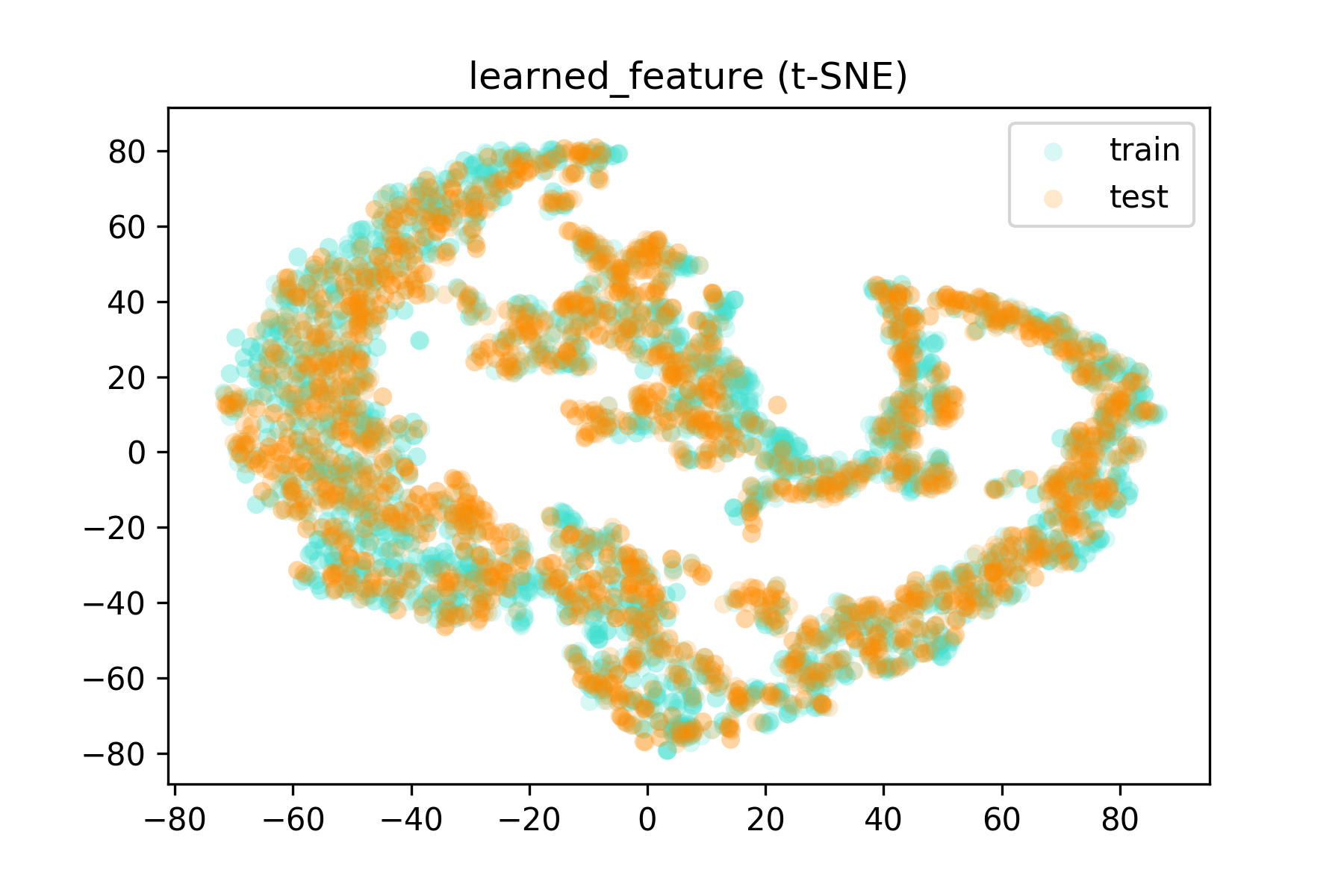}
		\caption{Online data wrt online model}
	\end{subfigure}\\
	\begin{subfigure}{0.23\textwidth}
		\centering
		\includegraphics[trim={1.cm 0.5cm 1.5cm 1.2cm},clip,width=\textwidth]{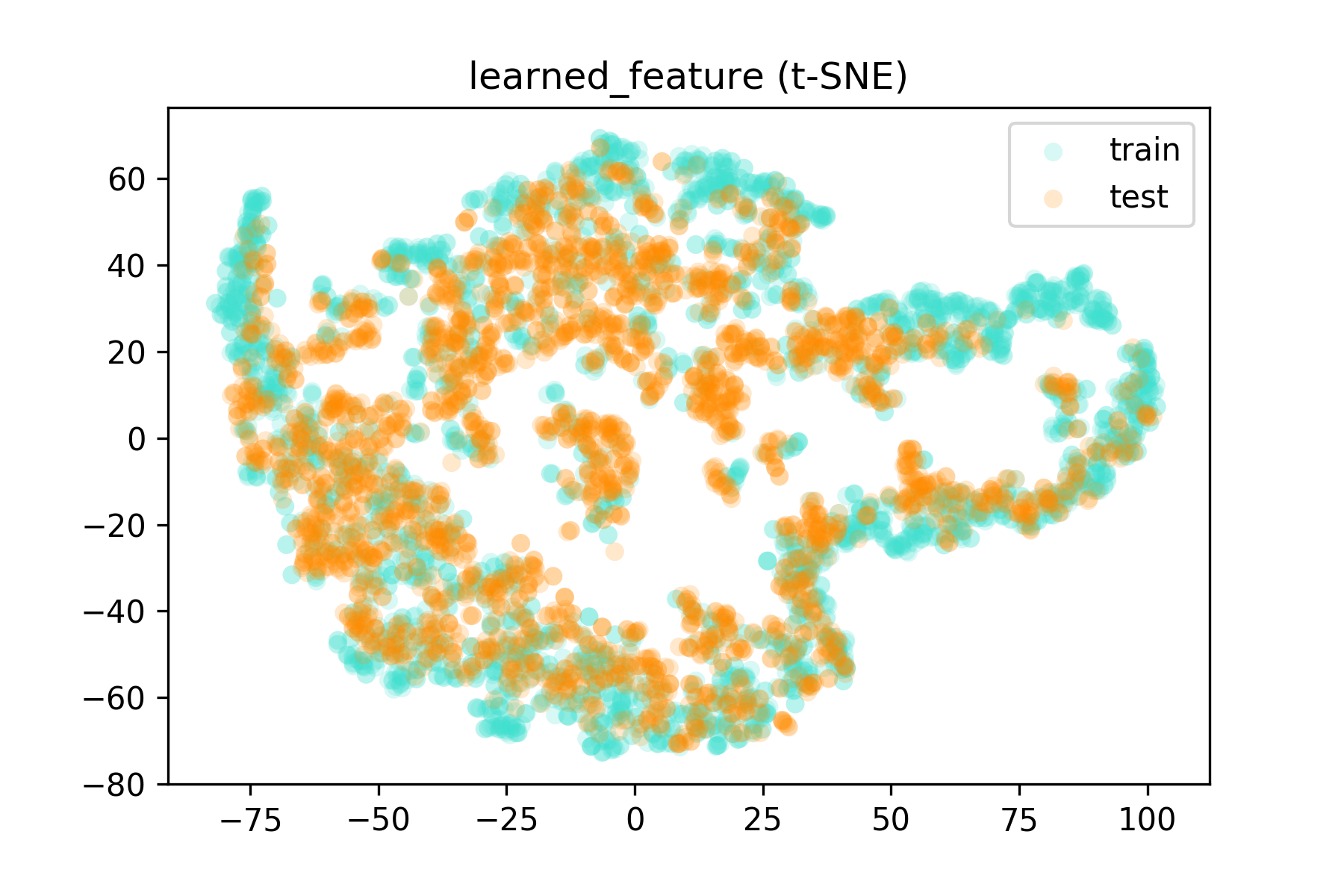}
		\caption{Batch data wrt batch model}
	\end{subfigure}
	\begin{subfigure}{0.23\textwidth}
		\centering
		\includegraphics[trim={1.cm 0.5cm 1.5cm 1.2cm},clip,width=\textwidth]{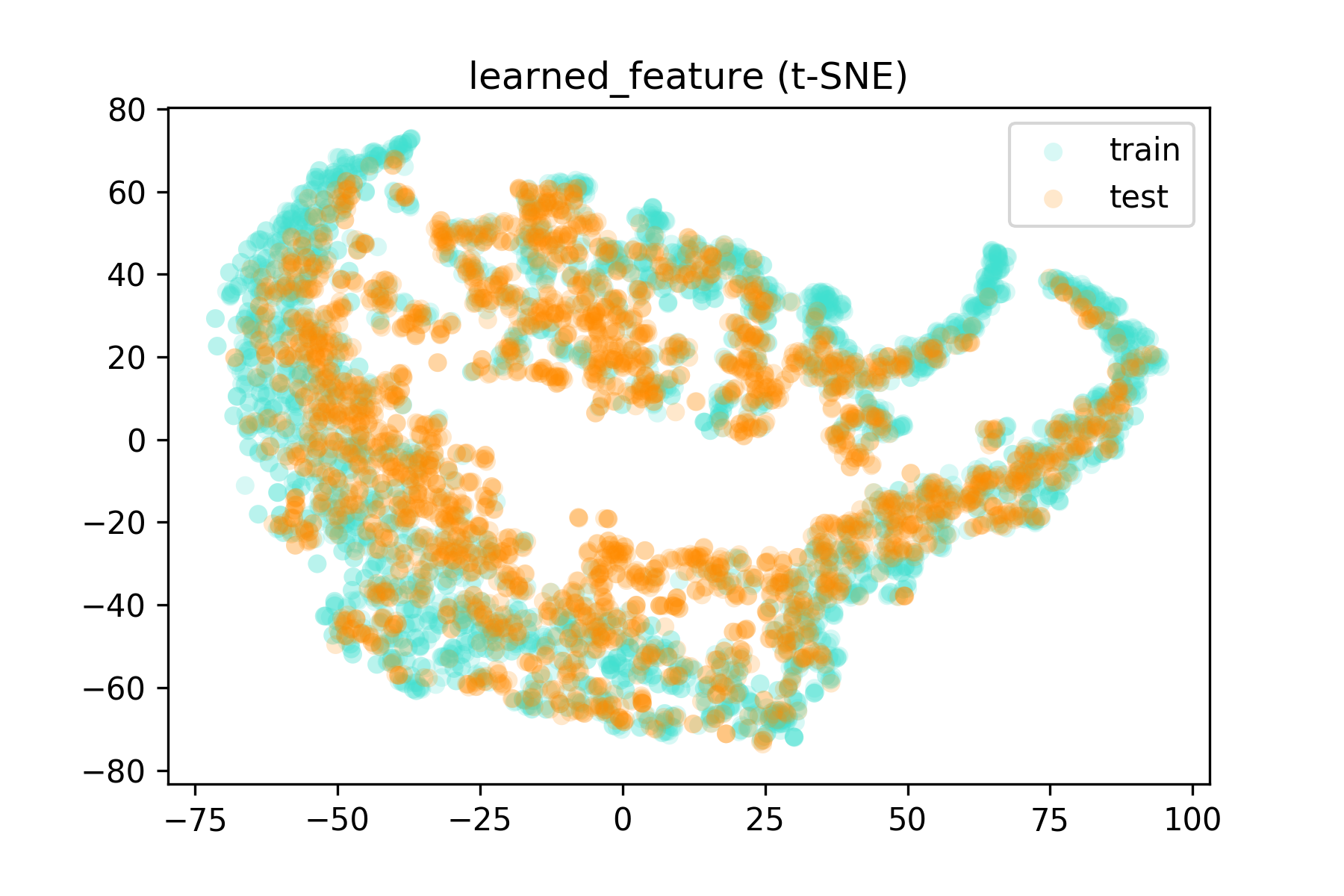}
		\caption{Online data wrt batch model}
	\end{subfigure}	
	\caption{The distributions (t-SNE) of the learned DNN feature in the last fully-connected layer (details are in Section~\ref{sec:generalizability}).} 
	\label{fig:feature distribution}
	\vspace{-3mm}
\end{figure}

To further analyze the difference between the DNNs trained using online and batch IL,
we embed the data in a two-dimensional space using t-Distributed Stochastic Neighbor Embedding (t-SNE)~\citep{maaten2008visualizing}, as shown in Fig.~\ref{fig:raw data distributions} and Fig.~\ref{fig:feature distribution}.
These figures visualize the data in both batch and online IL settings, where ``train'' denotes the data collected to train the policies and ``test'' denotes the data collected to evaluate the performance of the final policies after the learning phase.\footnote{For the online setting, the train data include the data in all DAgger iterations; for the batch setting, the train data include the same amount of data 
but collected by the expert policy. The figures plot a subset of 3,000 points from each data set.}

We first observe in Fig.~\ref{fig:raw data distributions} that, while the wheel speed data have similar training and testing distributions, the image distributions are fairly misaligned. The raw images are subject to changing lighting conditions, as the policies were executed at different times and days, and to various trajectories the robot stochastically traveled. Therefore, while the task (driving fast in the same direction) is \textit{seemingly} monotone, it actually is not. More importantly, the training and testing images were collected by executing different policies, which leads to different distributions of the neural networks inputs. This is known as the \textit{covariate shift} problem~\cite{shimodaira2000improving}, 
which can significantly complicate the learning process.

The policy trained with online IL yet still demonstrated great performance in the experiments. 
To further understand how it could generalize across different image distributions, we embed its feature distribution in Fig.~\ref{fig:feature distribution} (a) and (b).\footnote{The feature here are the last hidden layer of the neural network. The output layer is a linear function of the features.} 
Interestingly, despite the difference in the raw feature distributions in Fig.~\ref{fig:raw data distributions} (a) and (b), the DNN policy trained with online IL are able to map the train and test data to similar feature distributions, so that a linear combination (the last layer) of those features  is sufficient to represent a good policy. On the contrary, the DNN policy trained with batch IL fails to learn a coherent feature embedding, as shown in Fig.~\ref{fig:feature distribution} (c) and (d). This could explain the inferior performance of batch IL, and its inability to deal with the corner case in Fig.~\ref{fig:traj} (b). This evidence shows that our online learning system can alleviate the covariate shift issue caused by executing different policies at training and testing time.

\subsection{The Neural Network Policy}

\begin{figure}
	\centering
	\begin{subfigure}{0.23\textwidth}
		\centering
		\includegraphics[trim={2cm 3cm 2cm 2cm},clip,width=\textwidth]{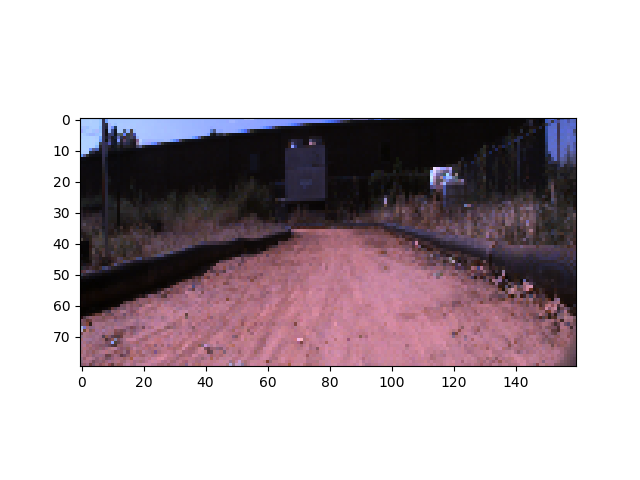}
		\caption{raw image}
	\end{subfigure}
	\begin{subfigure}{0.23\textwidth}
		\centering
		\includegraphics[trim={2cm 3cm 2cm 2cm},clip,width=\textwidth]{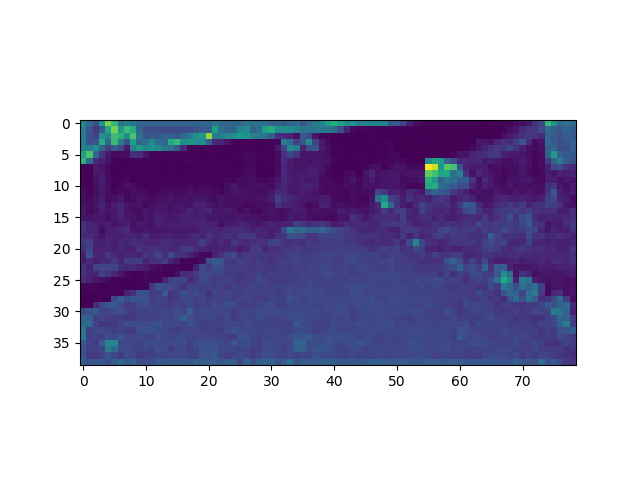}
		\caption{max-pooling1}
	\end{subfigure}\\
	\begin{subfigure}{0.23\textwidth}
		\centering
		\includegraphics[trim={2cm 3cm 2cm 2cm},clip,width=\textwidth]{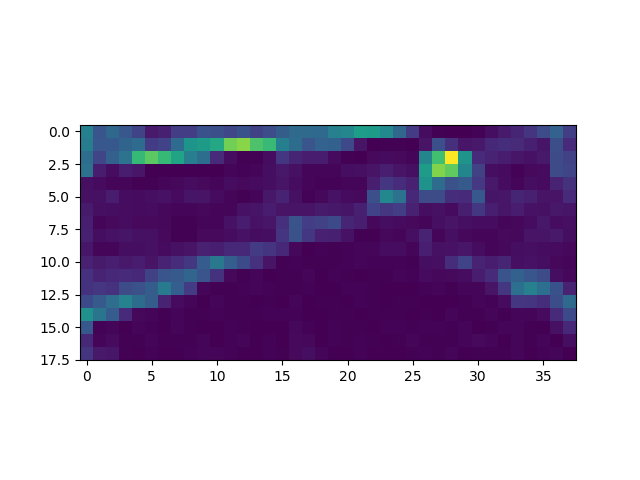}
		\caption{max-pooling2}\vspace{-1mm}
	\end{subfigure}
	\begin{subfigure}{0.23\textwidth}
		\centering
		\includegraphics[trim={2cm 3cm 2cm 2cm},clip,width=\textwidth]{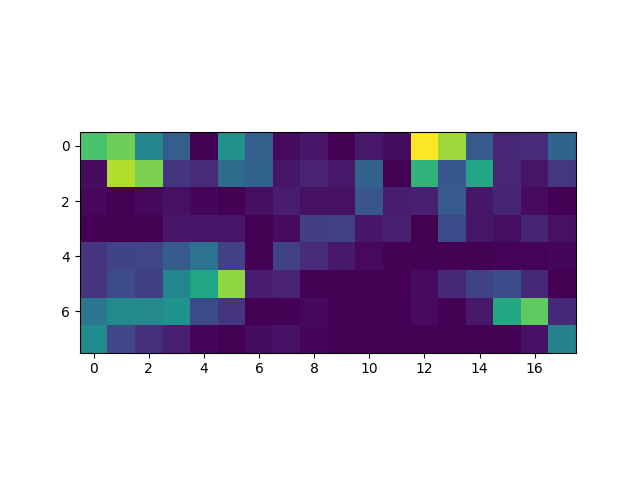}
		\caption{max-pooling3}\vspace{-1mm}
	\end{subfigure}
	\caption{The input RGB image and the averaged feature maps for each max-pooling layer.}
	\label{fig:features}
	\vspace{-3mm}
\end{figure}

\begin{figure}
	\centering
	\begin{subfigure}[b]{0.30\textwidth}
		\includegraphics[width=\textwidth]{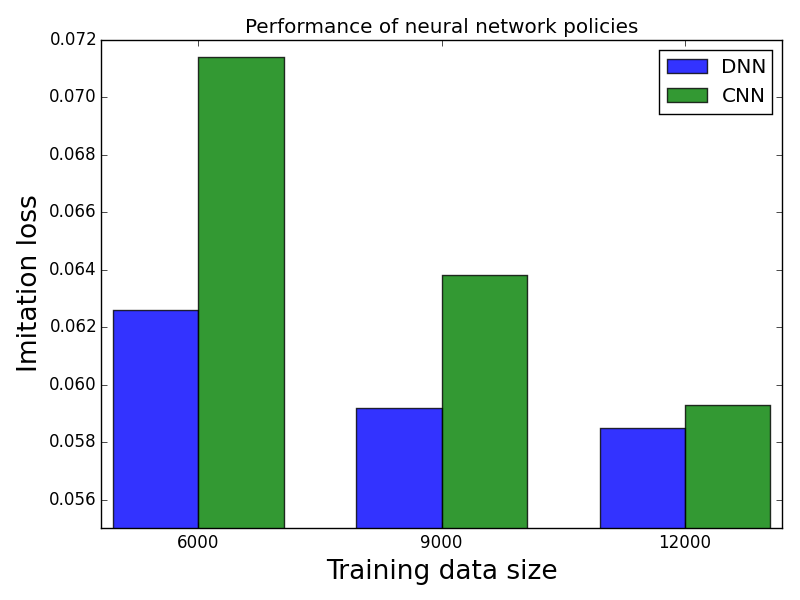}
	\end{subfigure}
	\caption{Performance comparison between our DNN policy and its CNN sub-network in terms of batch IL loss, where the horizontal axis is the size of data used to train the neural network policies. }
	\label{fig:dnn_vs_cnn}
	\vspace{-4mm}
\end{figure}

Compared with hand-crafted feature extractors, one main advantage of a DNN policy is that it can learn to extract both low-level and high-level features of an image and automatically detect the parts that have greater influence on steering and throttle. We validate this idea by showing in Fig.~\ref{fig:features} the averaged feature map at each max-pooling layer (see Fig. \ref{fig:deep_autorally_net}), where each pixel represents the averaged unit activation across different filter outputs. We can observe that at a deeper level, the detected salient features are boundaries of the track and parts of a building. In contrast, grass and dirt contribute little.

We also analyze the importance of incorporating wheel speeds in our task. We compare the performance of the policy based on our DNN policy and a policy based on only the CNN subnetwork (without wheel-speed inputs) in batch IL. The data was collected in accordance with Section~\ref{sec:data_collection}. 
Fig.~\ref{fig:dnn_vs_cnn} shows the batch IL loss in~\eqref{eq:batch learning} of different network architectures.
The full DNN policy in Fig.~\ref{fig:deep_autorally_net} achieved better performance consistently. While images contain position and orientation information, it is insufficient to infer velocities, which are a part of the (hidden) vehicle state. Therefore, we conjecture state-of-the-art CNNs (e.g. \cite{bojarski2017explaining}) cannot be directly used to perform both lateral and longitudinal controls, as we do here. By contrast, while without a recurrent architecture, our DNN policy learned to combine wheel speeds in conjunction with CNN to infer hidden state and achieve better performance. 

\vspace{-1mm}
\section{Conclusion}\vspace{-1mm}
\label{sec:conclusion}
We introduce an end-to-end system to learn a deep neural network control policy for high-speed driving that maps raw on-board observations to steering and throttle commands by mimicking a model predictive controller.  
In real-world experiments, our system was able to perform fast off-road navigation autonomously using a low-cost monocular camera and wheel speed sensors. We also provide an analysis of both online and batch IL frameworks, both theoretically and empirically and show that  our system, when trained with online IL,  learns generalizable features that are more robust to covariate shift than features learned with batch IL.

\vspace{-1mm}
\section*{Acknowledgements}
This work was partially supported by NSF NRI Awards
1637758 and 1426945.


\bibliographystyle{plainnat}
\bibliography{references}

\clearpage 

\onecolumn

\section*{APPENDIX}
\newcommand{\rinf}{{\mathrm \inf}}
\newcommand{\rd}{{\mathrm d}}
\newcommand{\rp}{{\mathrm p}}
\newcommand{\rQ}{{\mathrm q}}
\newcommand{\rtr}{{\mathrm{tr}}}
\newcommand{\rT}{{\mathrm{T}}}
\newcommand{\va}{{\bf a}}
\newcommand{\vc}{{\bf c}}
\newcommand{\vx}{{\bf x}}
\newcommand{\vy}{{\bf y}}
\newcommand{\vz}{{\bf z}}
\newcommand{\vk}{{\bf k}}
\newcommand{\vp}{{\bf p}}
\newcommand{\dvx}{{\bf dx}}
\newcommand{\vdu}{{\bf du}}
\newcommand{\vdy}{{\bf dy}}
\newcommand{\vq}{{\bf q}}
\newcommand{\vd}{{\bf d}}
\newcommand{\vr}{{\bf r}}
\newcommand{\vs}{{\bf s}}
\newcommand{\vf}{{\bf f}}
\newcommand{\vfgp}{{\vf_{\mathbb{GP}}}}
\newcommand{\vm}{{\bf m}}
\newcommand{\ve}{{\bf e}}
\newcommand{\vg}{{\bf g}}
\newcommand{\vG}{{\bf G}}
\newcommand{\vu}{{\bf u}}
\newcommand{\vv}{{\bf v}}
\newcommand{\vI}{{\bf I}}
\newcommand{\vh}{{\bf h}}
\newcommand{\vl}{{\bf l}}
\newcommand{\vZ}{{\bf Z}}
\newcommand{\vdx}{{\bf dx}}
\newcommand{\vdw}{{\bf dw}}
\newcommand{\vw}{{\bf w}}
\newcommand{\vW}{{\bf W}}
\newcommand{\vR}{{\bf R}}
\newcommand{\vE}{{\bf E}}
\newcommand{\vL}{{\bf L}}
\newcommand{\vK}{{\bf K}}
\newcommand{\vJ}{{\bf J}}
\newcommand{\vF}{{\bf F}}
\newcommand{\vM}{{\bf M}}
\newcommand{\vD}{{\bf D}}
\newcommand{\vN}{{\bf N}}
\newcommand{\vV}{{\bf V}}
\newcommand{\vH}{{\bf H}}
\newcommand{\vT}{{\bf T}}
\newcommand{\vC}{{\bf C}}
\newcommand{\vO}{{\bf O}}
\newcommand{\vQ}{{\bf Q}}
\newcommand{\vB}{{\bf B}}
\newcommand{\vS}{{\bf S}}
\newcommand{\vX}{{\bf X}}
\newcommand{\vY}{{\bf Y}}
\newcommand{\vP}{{\bf P}}
\newcommand{\vA}{{\bf A}}
\newcommand{\fq}{\mathfrak{q}}
\newcommand{\Vx}{{V_{\vx}}}
\newcommand{\Vxx}{{V_{\vx\vx}}}
\newcommand{\pat}{{\partial_{t}}}
\newcommand{\pax}{{\partial_{\vx}}}
\newcommand{\paxx}{{\partial_{\vx\vx}}}
\newcommand{\psix}{{\Psi_{\vx}}}
\newcommand{\psixx}{{\Psi_{\vx\vx}}}
\newcommand{\tx}{{\tilde{\bf x}}}
\newcommand{\bu}{{\bar{\bf u}}}
\newcommand{\bx}{{\bar{\bf x}}}
\newcommand{\tdx}{{\rd\tilde{\bf x}}}
\newcommand{\tq}{{\tilde{ q}}}
\newcommand{\bq}{{\bar{ q}}}
\newcommand{\tX}{{\tilde{\bf X}}}
\newcommand{\vxi}{{\mbox{\boldmath$\xi$}}}
\newcommand{\vdtheta}{{\bf \delta \theta} }
\newcommand{\vde}{{\mbox{\boldmath$\delta$}}}
\newcommand{\vvpi}{{\mbox{\boldmath$\varpi$}}}
\newcommand{\vpi}{{\mbox{\boldmath$\pi$}}}
\newcommand{\vdomega}{{\bf d\omega}}
\newcommand{\vlambda}{{\mbox{\boldmath$\lambda$}}}
\newcommand{\VGamma}{{\mbox{\boldmath$\Gamma$}}}
\newcommand{\VTheta}{{\mbox{\boldmath$\Theta$}}}
\newcommand{\VPhi}{{\mbox{\boldmath$\Phi$}}}
\newcommand{\vPhi}{{\mbox{\boldmath$\Phi$}}}
\newcommand{\VOmega}{{\mbox{\boldmath$\Omega$}}}
\newcommand{\VY}{{\mbox{\boldmath$\Upsilon$}}}
\newcommand{\vphi}{{\mbox{\boldmath$\phi$}}}
\newcommand{\VPsi}{{\mbox{\boldmath$\Psi$}}}
\newcommand{\vepsilon}{{\mbox{\boldmath$\epsilon$}}}
\newcommand{\VSigma}{{\mbox{\boldmath$\Sigma$}}}
\newcommand{\valpha}{{\mbox{\boldmath$\alpha$}}}
\newcommand{\vmu}{{\mbox{\boldmath$\mu$}}}
\newcommand{\vSigma}{{\mbox{\boldmath$\Sigma$}}}
\newcommand{\vbeta}{{\mbox{\boldmath$\beta$}}}
\newcommand{\vomega}{{\mbox{\boldmath$\omega$}}}
\newcommand{\vtau}{{\mbox{\boldmath$\tau$}}}
\newcommand{\vdtau}{{\mbox{\boldmath$d\tau$}}}
\newcommand{\vtheta}{{\mbox{\boldmath$\theta$}}}\newcommand{\dataset}{{\cal D}}
\newcommand{\fracpartial}[2]{\frac{\partial #1}{\partial #2}}
\newcommand{\vcalS}{{\mbox{\boldmath$\cal{S}$}}}
\newcommand{\vcalU}{{\mbox{\boldmath$\cal{U}$}}}
\newcommand{\vcalD}{{\mbox{\boldmath$\cal{D}$}}}
\newcommand{\vcalJ}{{\mbox{\boldmath$\cal{J}$}}}
\newcommand{\vcalE}{{\mbox{\boldmath$\cal{E}$}}}
\newcommand{\vcalF}{{\mbox{\boldmath$\cal{F}$}}}
\newcommand{\vcalL}{{\mbox{\boldmath$\cal{L}$}}}
\newcommand{\vcalZ}{{\mbox{\boldmath$\cal{Z}$}}}
\newcommand{\vcalG}{{\mbox{\boldmath$\cal{G}$}}}
\newcommand{\vcalN}{{\mbox{\boldmath$\cal{N}$}}}
\newcommand{\vcalM}{{\mbox{\boldmath$\cal{M}$}}}
\newcommand{\vcalH}{{\mbox{\boldmath$\cal{H}$}}}
\newcommand{\vcalC}{{\mbox{\boldmath$\cal{C}$}}}
\newcommand{\vcalO}{{\mbox{\boldmath$\cal{O}$}}}
\newcommand{\vcalP}{{\mbox{\boldmath$\cal{P}$}}}
\newcommand{\vcalB}{{\mbox{\boldmath$\cal{B}$}}}
\newcommand{\vcalA}{{\mbox{\boldmath$\cal{A}$}}}
\newcommand{\vcalg}{{\mbox{\boldmath$\cal{g}$}}}
\newcommand{\T}{^\mathsf{T}}
\newcommand{\mX}{\mathcal{X}}
\newcommand{\mF}{\mathcal{F}}
\newcommand{\tmu}{\tilde{\bm{\mu}}}
\newcommand{\tf}{{\tilde{\bf \vf}}}
\newcommand{\tdmu}{{\rd\tilde{\bf \vmu}}}
\newcommand{\tPsi}{{\tilde{\bf \Psi}}}
\newcommand{\bPsi}{{\bar{ \Psi}}}
\newcommand{\tSigma}{{\tilde{\bf \vSigma}}}
\newcommand{\tdSigma}{{\rd\tilde{\bf \vSigma}}}
\hyphenation{op-tical net-works semi-conduc-tor}
\newtheorem{mytheo}{Theorem}
\newtheorem{mydef}{Definition}

\subsection{Introduction}
In this supplementary material we provide details of the cost function and  model predictive control (MPC) expert used for learning the neural network policies in the main paper.

\subsection{Task cost function} 
The position cost $\text{cost}_{\textbf{pos}}(s)$ for the high-speed navigation task is a 16-term cubic function of the vehicle's global position $(x,y)$:
\begin{align*}\label{eq:track_poly}
		\text{cost}_{\textbf{pos}}(s) = c_0 + c_1 y + c_2 y^2 + c_3 y^3 + c_4 x + c_5 x y \nonumber\\+ c_6 x y^2 + c_7 x y^3 + c_8 x^2 + c_9 x^2 y + c_{10} x^2 y^2 + c_{11} x^2 y^3  \\+ c_{12} x^3 + c_{13} x^3 y + c_{14} x^3 y^2 + c_{15} x^3 y^3. 
\end{align*}
The coefficients in this cost function were identified by performing a regression to fit the track's boundary: First, a thorough GPS survey of the track was taken. Points along the inner and the outer boundaries were assigned values of $-1$ and $+1$, respectively, resulting in a zero-cost path along the center of the track. The coefficient values $c_i$ were then determined by a least-squares regression of the polynomials in $\text{cost}_{\textbf{pos}}(s)$ to fit the boundary data.

The speed cost $cost_{\textbf{spd}} = \norm{v_x - v_{\textbf{desired}} }^2 $ is a quadratic function which penalizes the difference between the desired speed $ v_{\textbf{desired}}$ and the longitudinal velocity $v_x$ in the body frame. The side slip angle cost is defined as
$
\text{cost}_{\textbf{slip}}(s) = -\arctan^2(\frac{v_y}{\|v_x\|})
$, 
where $ v_y$ is the lateral velocity in the body frame. 
The action cost is a quadratic function defined as $\text{cost}_{\textbf{act}}(a) = \gamma_1 a_1 + \gamma_2 a_2
$, where  $a_1$ and $a_2$ correspond to the steering and the throttle commands, respectively. In the experiments, $\gamma_1=1$ and $\gamma_2=1$ were selected.

\subsection{Model Predictive Controller (MPC)}
In this work, we use a model predictive controller (MPC) as the expert for imitation learning. The MPC expert is based on 1)  Sparse Spectrum Gaussian Process (SSGP) dynamics model, and 2) Differential Dynamic Programming (DDP) trajectory optimizer. In this section we provide a description of these techniques.

\subsubsection{Sparse Spectral Gaussian processes dynamics model}
Operating a vehicle at high speed in a stochastic environment results in complex dynamics that cannot be represented well by physics-based models. In this work we consider a statistical model, Sparse Spectrum Gaussian Processes~\cite{quia2010sparse} (SSGPs), which can be learned from real data.

Consider the task of learning the function (e.g., state transition in our case) $f: \reals^d \to \reals$, given IID data $\DD=\{x_i, y_i\}_{i=1}^n$,  with each pair  related by 
\vspace{-0.1 cm}
\begin{equation} \label{eq_target_function}
y = f(x) + \epsilon, \quad \epsilon \sim \NN(0, \sigma_n^2),
\end{equation}
where $\epsilon$ is an independent additive Gaussian noise. 
Gaussian process regression (GPR) is a principled way of performing Bayesian inference in function space, assuming that function $f$ has a prior distribution 
$f \sim \mathcal{GP}(m, k)$,
with mean function $m: \reals^d \to \reals$ and kernel $k: \reals^d \times \reals^d \to \reals$. Without loss of generality,  we assume $m(x) = 0$. 
Exact GPR is challenging for large datasets due to its $O(n^3)$ time and $O(n^2)$ space complexity \cite{williams2006gaussian}, which is a direct consequence of having to store and invert an $n\times n$ Gram matrix. 
Based on Bochner's theorem, continuous shift-invariant kernels can be unbiasedly approximated by an explicit finite-dimensional feature map. Leveraging this approximation, we consider SSGPs which is a class of Gaussian processes with kernel in the form:
\begin{gather} 
  k(x,x') = \phi(x)^T\phi(x') + \sigma_n^2 \delta(x-x'),
  \; \phi(x) = \begin{bmatrix} \phi^c(x) \\ \phi^s(x) \end{bmatrix},  \\ \nonumber
\phi^c_i(x) = \sigma_k \cos(\omega_i^T x), \;
\phi^s_i(x) = \sigma_k \sin(\omega_i^T x), \;
\omega_i \sim p(\omega),
\end{gather}
where function $\phi: \reals^{d} \rightarrow \reals^{2m} $ is the explicit finite-dimensional feature map, scalar $\sigma_k$ is a scaling coefficient, function $\delta$ is the Kronecker delta function, and vectors $\omega_i$ are sampled according to the spectral density $p(\omega)$ of the kernel to approximate. 
Because of the explicit finite-dimensional feature map $\phi$,
each SSGP is equivalent to a Gaussian distribution over the weights of the features $w \in \reals^{2m}$.
Assume that prior distribution of weights $w$ is $\NN(0, I)$ and that the feature map is  fixed. Conditioned on data $\DD=\{x_i, y_i\}_{i=1}^n$, the posterior distribution of $w$ is
\begin{gather}
w \sim \NN(\alpha, \;\sigma_n^2A^{-1}),  \label{eq_SSGP_weight_dist}\\
\alpha = A^{-1}\Phi Y, \quad A = \Phi \Phi^T + \sigma_n^2 I,  \label{eq_SSGP_parts}
\end{gather}
which can be derived through Bayesian linear  regression.
In \eqref{eq_SSGP_parts}, the column vector $Y$ and the matrix $\Phi$ are specified by the data $\DD$: $Y = \begin{bmatrix}y_1 & \ldots & y_n \end{bmatrix}^T$,  
$\Phi = \begin{bmatrix} \phi(x_1)& \ldots &\phi(x_n) \end{bmatrix}$.
Consequently, the posterior distribution over the output $y$ in \eqref{eq_target_function} at a test point $x$ is \emph{exactly Gaussian}, in which the posterior variance explicitly captures the model uncertainty in predicting $f(x)$:
\begin{gather} \label{eq_ssgpr}
p(y|x) = \NN (\alpha^T \phi(x), \; \sigma^2_n + \sigma^2_n \|\phi(x)\|^2_{A^{-1}}).
\end{gather}

We consider multivariate outputs by utilizing conditionally independent scalar models for each output dimension, \ie, we assume that, for outputs in different dimension $y_a$ and $y_b$, $p(y_a, y_b|x) = p(y_a|x)p(y_b|x)$.  In addition, hyper-parameters $\sigma_n, \sigma_k$ are optimized maximizing the GP marginal likelihood \cite{williams2006gaussian}.

Given samples from the state space dynamics\footnote{The samples can be collected by driving the vehicle manually or using a baseline controller.},  
\begin{align} 
x_{k+1} = x_k + f(x_{k}, u_{k}) + w_{k}, ~~~&w_{k} \sim \NN(0, \Sigma_w), \label{eq_dynamics}
\end{align}
We create a SSGP model of $f$. Given a distribution of the current state $p(x_{k})=\NN(\mu_{k}, \Sigma_{k})$ , we can compute the predictive distribution of the next state $p(x_{k+1}) \approx \NN(\mu_{k+1}, \Sigma_{k+1})$ by linearizing the predictive mean function \citep{pmlr-v70-pan17a}. The resulting approximate predictive distribution can be represented as follows: 
\begin{equation}\label{gpdyn}
\begin{split}
\mu_{k+1} &= \mu_k + \E f_k \\
 \Sigma_{k+1} &= \Sigma_k + \Cov f_k + \Cov(x_k ,f_k) + \Cov(f_k,x_k ) .
\end{split}
\end{equation} 
See \cite{pmlr-v70-pan17a} for the closed-form expressions of $\E f_k, \Cov f_k, \Cov(x_k ,f_k) $. Note that we use subscript $k$ to denote time step. Since the approximate predictive distribution is Gaussian, we define the belief of the dynamics at state $x_k$ as  $b_k=[\mu_k~ \text{vec}(\Sigma_k)]^{\rT}$, where $\text{vec}(\Sigma_k) $ is the vectorization of $\Sigma_k$. Therefore \eqref{gpdyn} can be written in a compact form
\begin{equation}\label{eq:belief_dyn}
b_{k+1} = \mF(b_k,u_k),  
\end{equation}
where $\mF$ is defined by (\ref{gpdyn}). The above equation corresponds to the belief space representation of the unknown dynamics (\ref{gpdyn}) in discrete-time.

\subsubsection{Differential Dynamic Programming}\label{sec_belief_dyn}
In order to incorporate dynamics model uncertainty explicitly, we perform trajectory optimization in the \textit{Gaussain belief space}.  Our proposed framework is based on Differential Dynamic Programming (DDP) \cite{tassa2008receding} \footnote{Since we use linear approximation of the dynamics, DDP and iterative LQR are interchangeable.},  at each iteration we create  a local model along a nominal trajectory through the belief space $(\bar{b}_k,\bar{u}_k) $ including: 1) a linear approximation of the belief dynamics model; 2) a second-order local approximation of the  value function. We denote the belief and control nominal trajectory as ($\bar{b}_{1:N},\bar{u}_{1:N}$) and deviations from this trajectory as $\delta b_k=b_k-\bar{b}_k$, $\delta u_k=u_k-\bar{u}_k$.  The  linear approximation of the belief dynamics along the nominal trajectory is
\begin{align}\label{belief_dyn}
\delta b_{k+1} \approx \left[ \begin{array}{cc}
\frac{\partial \mu_{k+1}}{\partial \mu_k} & \frac{\partial \mu_{k+1}}{\partial \Sigma_k} \\
\frac{\partial \Sigma_{k+1}}{\partial \mu_k}     &   \frac{\Sigma_{k+1}}{\partial \Sigma_k} \end{array} \right]\delta b_k +  \left[ \begin{array}{c}  \frac{\partial \mu_{k+1}}{\partial u_k} \\
\frac{\partial \Sigma_{k+1}}{\partial u_k}\end{array} \right]\delta u_k \eqqcolon \mF^b_k \delta b_k + \mF^u_k \delta u_k .
\end{align}
For a general non-quadratic cost function, we approximate it as a quadratic function along the nominal belief and control trajectory $(\bar{b}_{1:N},\bar{u}_{1:N})$, i.e., 
\begin{align}\label{eq_quadcost}
\LL(b_k,u_k) \approx\LL^0_k + (\LL^b_k)\T\delta b_k +  (\LL^u_k)\T\delta u_k +  \frac{1}{2}\left[ \begin{array}{c} \delta b_k \\ \delta u_k \end{array}\right]^{\rT} \left[\begin{array}{cc}\LL^{bb}_k & \LL^{bu}_k\\ \LL^{ub}_k & \LL^{uu}_k \end{array}\right] \left[ \begin{array}{c} \delta b_k \\ \delta u_k \end{array}\right],
\end{align}  
where superscripts denote partial derivatives, e.g., $\LL^b_k=\nabla_b\LL_k(b_k,u_k)$ and $\LL^0_k=\LL(b_k,u_k)$. We will use this superscript rule for dynamics and cost-related terms.
All partial derivatives  are computed analytically. 
Based on the dynamic programming principle, the value function is the solution to the Bellman equation

 \begin{align}\label{eq_bellman}
V(b_k,k) = \min_{u_k}\big(\underbrace{\mathcal{L}(b_k,u_k) + V\big(\mF(b_k,u_k),k+1 \big)}_{Q(b_k,u_k)}\big).
\end{align}
where  $V$ is the value function for the belief $b_k$ at time step $k$. At the terminal time step $V(b_N,N)=\E h(x(N))$ where $h(x(N))$ is the final cost, and  $\mathcal{L}(b_k,u_k)=\E l(x_k,u_k)$ with $l$ the running cost function.
Given the state dynamics in (\ref{eq_dynamics}) and  the cost in (\ref{eq_quadcost}),  a quadratic approximation of the value function along the nominal trajectory $\bar{b}_{1:N}$ can be obtained. We write this second order approximation as 
\begin{align}\label{eq_V_quad}
	V(b_k,k)\approx V^0_k + (V_k^b)\T\delta b_k + \frac{1}{2}\delta b_k\T V^{bb}_k\delta b_k. 
\end{align}
where again the superscripts denote partial derivatives, e.g., $V^b_k=\nabla_b V_k(b_k,u_k)$ and $V^0_k=V(b_k,u_k)$.
The coefficients in~\eqref{eq_V_quad} can be derived by expanding the $Q$-function defined in (\ref{eq_bellman}) along $(\bar{b}_{1:N},\bar{u}_{1:N})$:
\begin{equation}\label{eq_Q}
\begin{split}
Q_k(b_k+\delta b_k,u_k+\delta u_k)\approx Q_k^0+Q^b_k\delta b_k+Q^u_k\delta u_k +\frac{1}{2}\left[ \begin{array}{c} \delta b_k \\ \delta u_k \end{array}\right]^{\rT} \left[\begin{array}{cc}Q^{bb}_k & Q^{bu}_k\\ Q^{ub}_k & Q^{uu}_k \end{array}\right] \left[ \begin{array}{c} \delta b_k \\ \delta u_k \end{array}\right],
\end{split}
\end{equation}
where 
\begin{align}\label{back_Q}
Q^{b}_k&=\mathcal{L}_k^b + V_k^b\mF^b_k,~~~Q^{u}_k=\mathcal{L}_k^u + V_k^b\mF^u_k,\nonumber\\
Q^{bb}_k&=\mathcal{L}_k^{bb} +(\mF^b_k)^{\rT}V^{bb}_k \mF^b_k,~~~Q^{ub}_k=\mathcal{L}_k^{ub} +(\mF^u_k)^{\rT}V^{bb}_k \mF^b_k, \nonumber\\
Q^{uu}_k&=\mathcal{L}_k^{uu} +(\mF^u_k)^{\rT}V^{bb}_k \mF^u_k.
\end{align}
The local optimal control law is obtained by minimizing the approximated $Q$ function
\begin{equation}\label{policy}
\begin{split}
\delta\hat{u}_k = \arg\min_{\delta u_k} \big[Q_k(b_k+\delta b_k,u_k+\delta u_k) \big] = -(Q^{uu}_k)^{-1}Q_k^u -(Q^{uu}_k)^{-1}Q_k^{ub}\delta b_k,
\end{split}
\end{equation}
The new optimal control is obtained as $\hat{u}_k=\bar{u}_k+\delta\hat{u}_k.$ Plugging the optimal control of (\ref{policy}) into the approximated Q-function given by (\ref{eq_Q}) results in the following backward propagation of the value function 
\begin{align}\label{back}
V_{k-1}=V_k-Q^u_k(Q^{uu}_k)^{-1}Q_k^u,~~~ 
V^b_{k-1} &= Q^{b}_k - Q^u_k(Q^{uu}_k)^{-1}Q_k^{ub},~~~~ V^{bb}_{k-1} = Q^{bb}_k - Q^{vu}_k(Q^{uu}_k)^{-1}Q_k^{ub}   . \nonumber
\end{align}\normalsize
 The optimized control policy $\hat{u}_{1:N}$ is applied to the belief dynamics \ref{eq:belief_dyn} to generate a new nominal trajectory in a forward pass. We keep optimizing the control policy using this backward-forward scheme iteratively until convergence.

\end{document}